\documentclass[final,5p,times,twocolumn]{elsarticle}
\usepackage[hidelinks]{hyperref}
\usepackage{amssymb}
\usepackage{amsmath,amssymb,amsfonts}
\usepackage{multirow}

\begin{document}

\begin{frontmatter}

\title{Building RadiologyNET: Unsupervised annotation of a large-scale multimodal medical database}

\author[RITEH]{Mateja Napravnik}
\author[RITEH,AIRI]{Franko Hržić}
\author[MUG]{Sebastian Tschauner}
\author[RITEH,AIRI]{Ivan Štajduhar\corref{cor1}}
\cortext[cor1]{Corresponding author: istajduh@riteh.hr, +385 51 651 448}

\affiliation[RITEH]{organization={University of Rijeka -- Faculty of Engineering},
            addressline={Vukovarska 58}, 
            city={Rijeka},
            postcode={51000}, 
            country={Croatia}}

\affiliation[AIRI]{organization={Center for Artificial Intelligence and Cybersecurity},
            addressline={Radmile Matejcic 2}, 
            city={Rijeka},
            postcode={51000}, 
            country={Croatia}}

\affiliation[MUG]{organization={Medical University of Graz},
            addressline={Neue Stiftingtalstraße 6}, 
            city={Graz},
            postcode={8010}, 
            country={Austria}}

\newcommand{\radiologynetSize}{1,337,926}

\begin{abstract}

\noindent \textit{Background and objective:} The usage of machine learning in medical diagnosis and treatment has witnessed significant growth in recent years through the development of computer-aided diagnosis systems that are often relying on annotated medical radiology images. However, the availability of large annotated image datasets remains a major obstacle since the process of annotation is time-consuming and costly. This paper explores how to automatically annotate a database of medical radiology images with regard to their semantic similarity.

\noindent \textit{Material and methods:} An automated, unsupervised approach is used to construct a large annotated dataset of medical radiology images originating from Clinical Hospital Centre Rijeka, Croatia, utilising multimodal sources, including images, DICOM metadata, and narrative diagnoses. Several appropriate feature extractors are tested for each of the data sources, and their utility is evaluated using k-means and k-medoids clustering on a representative data subset.

\noindent \textit{Results:} The optimal feature extractors are then integrated into a multimodal representation, which is then clustered to create an automated pipeline for labelling a precursor dataset of {\radiologynetSize} medical images into $50$ clusters of visually similar images. The quality of the clusters is assessed by examining their homogeneity and mutual information, taking into account the anatomical region and modality representation. 

\noindent \textit{Conclusion:} The results suggest that fusing the embeddings of all three data sources together works best for the task of unsupervised clustering of large-scale medical data, resulting in the most concise clusters. Hence, this work is the first step towards building a much larger and more fine-grained annotated dataset of medical radiology images.

\end{abstract}

\begin{keyword}
medical data annotation \sep feature extraction \sep feature fusion \sep big data \sep multimodal representation \sep unsupervised machine learning
\end{keyword}

\end{frontmatter}


\section{Introduction}\label{sec:intro}
In recent years, leveraging machine learning (ML) in medical diagnosis and treatment has skyrocketed. However, one of the major barriers in developing efficient medical models and computer-aided diagnosis (CAD) systems is the availability of annotated datasets. Such datasets have often been manually curated by specialists, which is time-consuming, expensive, and subject to the annotators' level of knowledge~\cite{Nagy:2022:grazped_dataset,Irvin:2019:CheXpert_dataset,rajpurkar:2018:mura_dataset}. In this paper, we explore an automated, unsupervised approach to building a large annotated medical dataset from multimodal sources, which includes images, image metadata and textual diagnoses. Here, we use the term annotate to describe the process of assigning labels to images based on their semantic similarity -- pairs of images that are more semantically similar should have the same label, whereas pairs of images being more semantically dissimilar should have different labels. 

Transfer learning (TL)~\cite{weiss2016survey,pan2010survey} is an ML approach that involves utilising knowledge gained from solving one problem to solve another similar problem. TL is an efficient way of modelling that requires less data and less time, compared to modelling from random weight initialisation. It involves training complex models with large amounts of readily available data and then fine-tuning the pretrained models to fit problem domains with smaller datasets. 
One of the most well-known data sources for building pretrained models for image processing is ImageNet~\cite{krizhevsky2017imagenet}, which consists of hierarchically-organised real-world photos. In a standard setup, these pretrained models become publicly available once they are perfected, which in turn enables other researchers to obtain better-performing models on alternate domains, and in a shorter time. ImageNet pretrained weights of contemporary model architectures are available for the most popular ML development platforms, such as PyTorch~\cite{paszke2019pytorch}, Tensorflow~\cite{abadi2016tensorflow}, etc.

TL from natural image datasets using standard architectures and corresponding pretrained weights has also become the norm for medical image processing using deep learning~\cite{raghu2019transfusion, Alzubaidi:2021:BigData}. However, when considering medical ML image processing, there is evidence that medical radiology datasets are more suitable as the source for learning pretrained models compared to natural image datasets, such as ImageNet~\cite{alzubaidi2020towards,mustafa2021supervised}.
This is understandable because of the semantic differences in images which are caused by a large distribution shift between the domains. Moreover, other properties of images between differing TL domains -- such as the number of channels and colour depth -- make TL more difficult.

To counter this problem, in this work, we explore the possibility of building a large annotated dataset of medical radiology images -- RadiologyNET -- to be used for learning the weights of contemporary model architectures, which could then be publicly shared over an online platform -- similar to ImageNet pretrained weights. Desired properties of RadiologyNET are: (1) cover major imaging modalities, examination protocols, and anatomical regions; and (2) use a large number of visually distinct classes (i.e.~fine grained image labelling). For this purpose, we utilise a large dataset of medical radiology images gathered retrospectively from the Clinical Hospital Centre (CHC) Rijeka's Picture Archiving and Communication System (PACS). Approval for conducting the research was obtained from the competent Ethics Committee. Image annotation is performed using unsupervised learning techniques from available multimodal data sources: images, DICOM tags and narrative diagnoses. We are not interested in the visual concept categories descriptions, only that the annotation has good clustering properties, grouping semantically similar images into the same clusters, while separating pairs of semantically dissimilar images.

The main contribution of this work is to explain the process used for annotating the RadiologyNET dataset to give validity to the TL pretrained models that we intend to build and publish in the future. Image annotation in this context is an evolving process and may result in future RadiologyNET refinements, henceforth resulting in better TL pretrained models. For example, one limitation of the current version of the RadiologyNET dataset is a relatively small number of distinct classes compared to the ImageNet database; this issue will be resolved in future dataset versions. Another contribution of this work is a comprehensive description of methods and approaches which should be considered when parsing large medical data repositories.

This paper is structured as follows. In section~\ref{sec:m&m}, we describe the characteristics of the individual data sources of our dataset, the utilised preprocessing and feature extraction methods, and the entire experimental setup. A more detailed description of the data extraction and preprocessing is available in the Appendix. The experiments were conducted on a smaller subset of the RadiologyNET dataset due to the computational constraints of the experimental pipeline. In section~\ref{sec:results}, we first present and compare the evaluation results of different feature extraction and clustering setups on this subset. After applying the best solution to the rest of the data, we describe the annotation characteristics of the current version of the RadiologyNET dataset in section~\ref{ssec:radiologynet_data_set}. Finally, in section~\ref{sec:con}, we discuss the reach and limitations of the annotated dataset and how it will be used in the future to build and share pretrained convolutional neural network models of varying architectures.

\section{Material and methods}\label{sec:m&m}
\newcommand{\numberOfImagesInSubset}{$135,775$}
\newcommand{\numberOfExamsInSubset}{$63,160$}
\newcommand{\numberOfImagesInEachModality}{$27,155$}
\newcommand{\maxNrOfImagesInExam}{$15$}
\newcommand{\distinctBPEcnt}{$29$}
\newcommand{\lowestNrOfCharactersInDiagnoses}{$5$}
\newcommand{\highestNrOfCharactersInDiagnoses}{$5312$}
\newcommand{\meanNrOfCharactersInDiagnoses}{$822$}
\newcommand{\meanNrOfSentencesInDiagnoses}{$10$}
\newcommand{\meanNrOfWordsInDiagnoses}{$80$}
\newcommand{\bpeFullPercent}{$40.6\%$}
\newcommand{\bpeEmptyPercent}{$59.4\%$}

The layout of this section is as follows. First, in section \ref{ssec:data_cleanup}, we describe the used dataset for each of the data sources separately. This involves the information on how raw data was processed to prepare it for subsequent steps. In section \ref{ssec:feature_extraction}, we explain the feature extraction techniques employed for each extracted data source and how they were implemented. Finally, in section \ref{ssec:experimental_setup}, we give an overview of the clustering methods and how the effectiveness of the resulting groups was measured.

\subsection{Experimentation dataset}\label{ssec:data_cleanup}

\newcommand{\StudyDescEmptyPercentage}{$7.39\%$}
\newcommand{\ProtocolNameEmptyPercentage}{$10.9\%$}
\newcommand{\ReqProcedureDescEmptyPercentage}{$33.8\%$}
\newcommand{\initialNrOfDcmTags}{$654$}
\newcommand{\dcmTagCardinalityThreshold}{50}
\newcommand{\dcmTagFillrateThreshold}{$35\%$}
\newcommand{\dcmTagUniqueValuesThreshold}{$2$}
\newcommand{\dcmTagNrOfParsedVars}{$55$}
\newcommand{\dcmTagNrOfParsedCategoricalVars}{$27$}
\newcommand{\dcmTagNrOfParsedContinuousVars}{$28$}
\newcommand{\nrOfRegexWritten}{$53$}
\newcommand{\originalEncodedDfSize}{321}

From the original dataset, which is described in detail in~\ref{sec:datpp}, we sampled a subset of {\numberOfImagesInSubset} DICOM files and adjoined textual diagnoses. The subset was balanced with regard to imaging modality; therefore each modality had an equally large representation of {\numberOfImagesInEachModality} instances. This was done to ensure that our findings equally apply to any of the most commonly occurring imaging modalities. The {\numberOfImagesInSubset} sampled DICOM files were linked to {\numberOfExamsInSubset} different medical examinations, meaning there were {\numberOfExamsInSubset} distinct diagnoses in the subset, and each exam had between $1$ and {\maxNrOfImagesInExam} images in the subset.
The dataset was randomly split into the train, test and validation subsets: $80\%$ of all exams were used for training, $10\%$ was used for testing and the remainder for validation. The exact subset sizes are provided in Table~\ref{tab:train_test_val_split}.

\begin{table}[!tb]
\centering
\caption{The sizes of train, test and validation subsets.}
\label{tab:train_test_val_split}
\footnotesize
\begin{tabular}{lrr} 
\hline\hline
\multicolumn{1}{c}{\textbf{Subset}} & \multicolumn{1}{c}{\begin{tabular}[c]{@{}c@{}}\textbf{Exam (diagnoses)}\\\textbf{ count}\end{tabular}} & \multicolumn{1}{c}{\begin{tabular}[c]{@{}c@{}}\textbf{DICOM file}\\\textbf{ count}\end{tabular}}  \\ 
\hline
Train   & 50,528 (80.00\%)     & 108,542 (79.94\%)    \\ 

Test    & 6,316 (10.00\%)    & 13,637 (10.05\%)  \\ 

Validation & 6,316 (10.00\%)  & 13,596 (10.01\%)     \\ 
\hline
\textbf{Total}  & 63,160 (100.00\%)    & 135,775 (100.00\%)                        \\
\hline\hline
\end{tabular}
\end{table}

\begin{figure*}[!tb]
    \centering
    \includegraphics[height=350px]{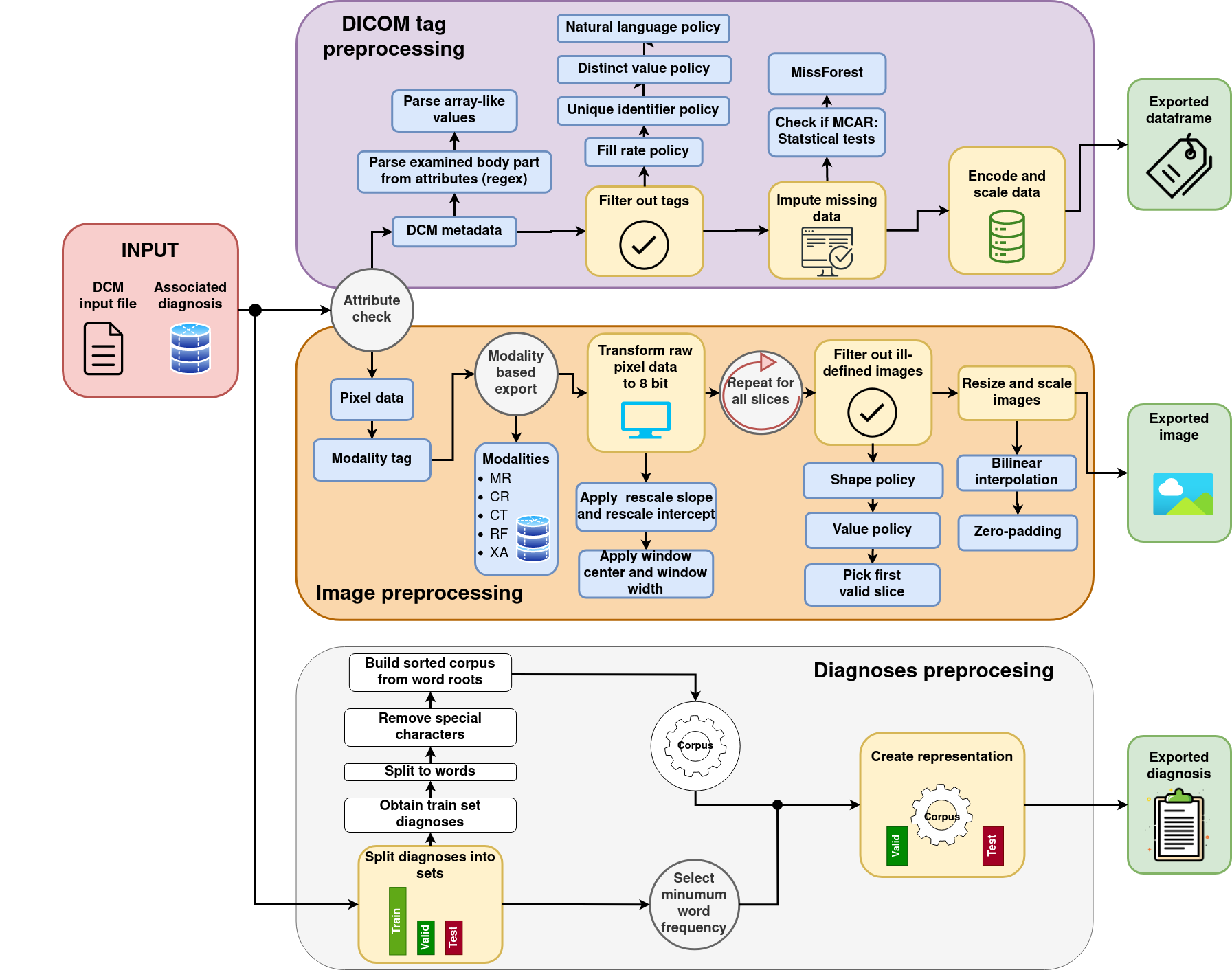}
    \caption{A graphical depiction of the process utilised for exporting images, DICOM tags metadata, and narrative diagnoses. Each DICOM file has an associated diagnosis. Each of the three data sources required a distinct preprocessing approach, and each approach had its peculiarities. For example, DICOM tags require additional filtering and the filling in of missing values. On the other hand, images come in different modalities, each requiring a specific approach. Moreover, images are often stored as 12-16 bits arrays, while the monitors and many algorithms support only 8-bit data. Consequently, the images required conversion and additional scaling and resizing. On the other hand, textual diagnoses are written in narrative form. This means that the adverbs, nouns, and verbs had to be stripped to their roots. Moreover, to create a representation of the diagnosis, it was necessary for many methods to build a corpus based on the frequent words in the training set (Table.~\ref{tab:train_test_val_split}). 
    All details concerning data parsing are presented in section~\ref{ssec:data_cleanup} and in the Appendix. The resulting extracted data is then fed to the feature extraction models.}
    \label{fig:export_image_and_tags}
\end{figure*}

Each DICOM file consists of two main parts: the raw image and the metadata describing the image (DICOM tags). Moreover, each DICOM file is accompanied by a narrative diagnosis. All three data sources (i.e. image, tags and diagnosis) were processed independently of each other. The data extraction process is illustrated in Fig.~\ref{fig:export_image_and_tags}.
In the remainder of the text, we refer to one recorded instantiation of the three sources' values -- or a part of it -- a data point, or an instance.

In subsequent sections, we describe the preprocessing steps that were taken for the three data sources to adapt each of them for feature extraction. Because of the overwhelming complexity of these steps, technical details were transferred to the Appendix and omitted from the main body to make the text easier to read. Moreover, if the reader is not interested in the technical details concerning data preprocessing altogether, we recommend skipping ahead to section~\ref{ssec:feature_extraction}. All the choices concerning data preprocessing and feature extraction were determined using training and validation subsets.

\subsubsection{DICOM tags}\label{sec:dcmtag}
The available DICOM tags were analysed, which was followed by identifying a few problems: finding useful DICOM tags, parsing tags with multiple values, and handling missing data, with the latter being the most prominent issue. Dropping features with missing data can lead to the loss of valuable information, while inadequate handling of missing data can lead to confounded results \cite{enders2022applied-missing-data}. The steps involved in tackling each of these problems are outlined in the next paragraph, while an in-depth description is given in \ref{AP:DICOM TAGS}. The entire process is also illustrated in  Fig.~\ref{fig:export_image_and_tags}.

\begin{figure}[!tb]
    \centering
    \includegraphics[width=\linewidth]{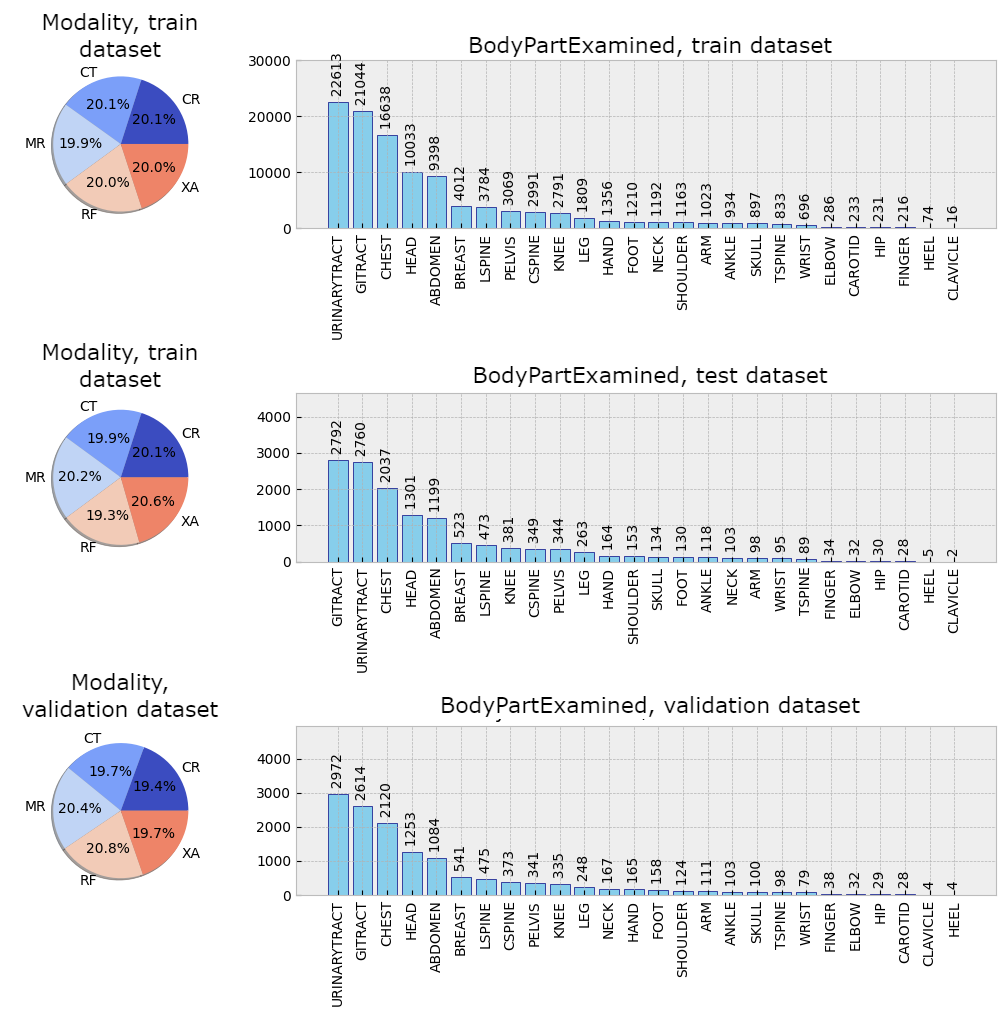}
    \caption{The distribution of image modalities (left) and examined body parts (right) in each subset of the used dataset is approximately equal.}
    \label{fig:mod_bpe_hist_by_dataset}
\end{figure}

We intended to use the \textit{BodyPartExamined} (BPE) tag (which describes the anatomical region shown in the image) for evaluation purposes. Therefore, as the first step, (1) it was deemed crucial to reconstruct missing BPE values as accurately as possible. BPE was empty in {\bpeEmptyPercent} cases, but this was alleviated by analysing other available DICOM tags. Namely, under advice from the radiologist, a set of regular expressions were written based on which BPE values could be inferred from other DICOM tags.  The final distribution of examined body parts across all datasets, coupled with the modality distribution, is shown in Fig.~\ref{fig:mod_bpe_hist_by_dataset}.
Secondly, (2) some DICOM tags contained array-like values. These were parsed so that each array-like tag was split into multiple single-value tags. After this, (3) filtering of DICOM tags was performed to determine which tags could be of use. A fill-rate threshold was imposed on each tag, and each tag which was non-empty in less than {\dcmTagFillrateThreshold} cases was excluded from further use. Any tags containing unique identifiers, natural language, or which had less than two distinct values were dropped as well. Furthermore, (4) analysis of data missingness was performed \cite{enders2022applied-missing-data, bhaskaran2014differenceMCAR} followed by imputation of missing data using MissForest \cite{Emmanuel2021missing_data_ml, stekhoven2012missforest, Tang2017RF_Missingdata}. Finally, (5) categorical variables were one-hot encoded and continuous variables were scaled to fit the range $[0.00, 1.00]$.

\subsubsection{Images}\label{sec:imgs}
\newcommand{\totalNrOfPNGs}{\numberOfImagesInSubset}
\newcommand{\imageTrainSubsetSize}{$108,542$}

Image preprocessing solves the problem of inconsistent image sizes and pixel depths.
Standardising these characteristics is crucial for every ML model to learn effectively, leading to more accurate and consistent results.
Several modalities were represented in the provided dataset. Every modality had its own peculiarities that need special addressing when extracting images from the raw images stored in DICOM files. The complete process of image extraction is depicted in Fig.~\ref{fig:export_image_and_tags}.

The raw images found in DICOM files can have a depth between 12 and 16 bits~\cite{larobina2014medical}. Before applying feature extractors to images, their pixel depth should be unified, and pixel values normalised. Hence, it is preferable to transform raw images into 8-bit images and in a way that causes the least amount of information loss. Namely, the majority of computer displays used for commercial purposes are limited to displaying 8-bit images. Also, most of the ML algorithms are designed to process images where the intensity of pixels falls within the 8-bit spectrum. To achieve this, there are parameters found in DICOM metadata which can be used to appropriately export a DICOM image to an 8-bit image. 

The export process requires several parameters stored as DICOM tags. During the data sampling process, we initially verified if all the necessary parameters were available in the DICOM files to determine that the observed DICOM file was eligible for utilisation. 
The comprehensive procedure for converting raw DICOM pixel data to 8-bit images is outlined in \ref{AP:DICOM EXPORT}. In summary, the image export process consisted of the following: (1) tags \textit{RescaleIntercept} and \textit{RescaleSlope} were read out from DICOM metadata \cite{dicom_standard} and applied to the raw image; (2) tags \textit{WindowCenter} and \textit{WindowWidth} from DICOM metadata were used to transform the rescaled pixel values into 8-bit data; (3) a \textit{value} policy was implemented to filter out images which were completely monochrome (e.g.~black or grey); (4) all images whose shape was erroneous (for example, images stored as 1D vectors) were excluded by applying the \textit{shape} policy; finally, (5) each image was resized to $128\times128$ pixels using bilinear interpolation, and zero-padding was added where necessary, to preserve aspect ratio. The image size $128 \times 128$ was chosen after careful consideration of the size of the image dataset and limited available processing resources. Additionally, at $128 \times 128$, the level of detail preserved in the images was deemed sufficient to adequately compare their visual similarities. This hypothesis was confirmed by performing a small-scale experiment where the images were resized to $256\times256$, but the results showed no significant difference.

\subsubsection{Diagnoses}
Narrative diagnoses are written in Croatian language and contain information about diseases and patients' conditions. The Croatian language has several unique aspects concerning word forms, such as seven grammatical cases and different verb suffixes. 
Two nouns can have the same meaning but are typed differently if they are in different grammatical cases. Hence, the first step was to strip the words to their roots to capture their core meaning better. Namely, we were interested in grouping the diagnoses based on their word meaning similarities, as encoding entire sentences \cite{Yang2019sentence_encoders} while keeping words intact fell out of the scope of this paper. 
Although we are aware of vastly popular generative pretrained transformers (GPT), for this particular problem, we opted for computationally less demanding models with the open possibility to expand to the GPTs in the future work~\cite{openai:2023:gpt4}.

First, (1) all diagnoses from the training subset were split into separate words, with special characters (commas, semicolons, colons...) removed. Then, (2), the words were reduced to their roots using the Croatian stemmer published by Ljubešić et al.~\cite{ljubevsic:2007:stemmer}. (3) In the training subset, there were a total of $54,790$ distinct words from which a word corpus was built. This number encompasses words that appear at least once in the training set's diagnoses. It was often the case that words present in a small number of instances are actually a result of typographical errors (anomalies) made by physicians who manually write the diagnoses. Hence, in order to enhance the models' generalisation capabilities, (4) a parameter that regulates the least amount of word occurrences needed for a word to be included in the corpus was introduced.
Finally, the pipeline of preprocessing narrative diagnoses is illustrated in Fig.~\ref{fig:export_image_and_tags}.

\subsection{Feature extraction}\label{ssec:feature_extraction}

\begin{table*}[!tb]
\centering
\footnotesize
\caption{Explored hyperparameter value ranges for DICOM tags, image and diagnosis feature extraction. The values were originally taken from related work, and then refined empirically.}
\label{tab:table_of_all_tested_hyperparams_for_all_sources}

\begin{tabular}{p{1.7cm}p{2.7cm}p{5.2cm}p{6.4cm}}
\hline\hline
\textbf{Data Source} & \textbf{Feature extractor(s)} & \textbf{Hyperparameter} & \textbf{Tested values} \\ \hline

\multirow{7}{*}{DICOM tags}  & \multirow{5}{*}{AE} & Learning rate & $10^{-2}, 10^{-3}, 10^{-4}, 10^{-5}$ \\ \cline{3-4} 
 &  & Bottleneck size (embedding size) & 32, 64, 50, 75, 100 \\ \cline{3-4} 
 &  & Layer sizes & \begin{tabular}[c]{@{}l@{}}$1^{st}$: 300, 250, 200, 512, 256\\ $2^{nd}$: 200, 150, 125, 100, 256, 128\\ $3^{rd}$: 150, 125, 100, 75, 128, 64\end{tabular} \\ \cline{2-4} 
 & \multirow{2}{*}{PCA} & Solver & LAPACK~\cite{anderson1999lapack}, 
 ARPACK~\cite{lehoucq1998arpack}, randomised~\cite{martinsson2011pcarandomized} \\ \cline{3-4} 
 &  & Number of components (embedding size) & 32, 64, 50, 75, 100 \\ \hline

\multirow{3}{*}{ Images} &  \multirow{3}{*}{\begin{tabular}[c]{@{}l@{}}CAE, U-Net, \\ AttU-Net, R2U-Net\end{tabular}} & Learning rate & $10^{-4}, 10^{-5}, 10^{-6}, 10^{-7}$ \\ \cline{3-4} 
 &  & PCA solver & [Not applied], LAPACK, ARPACK, randomised \\ \cline{3-4} 
 &  & PCA number of components (embedding size) & [Not applied], 100, 500, 1000, 10000 \\

\hline

\multirow{4}{*}{ Diagnoses} & \begin{tabular}[c]{@{}l@{}}BOW, TF-IDF,\\ PV-DM, PV-DBOW\end{tabular} & \begin{tabular}[c]{@{}l@{}} Minimum word frequency \end{tabular} & \begin{tabular}[c]{@{}l@{}}5, 10, 50, 100, 500, 1000, 2000, 3500, 5000, 10000\end{tabular} \\ \cline{2-4}

& \multirow{3}{*}{\begin{tabular}[c]{@{}l@{}}PV-DM, PV-DBOW\end{tabular}} & Embedding size & 10, 25, 50, 100, 250, 500, 1000 \\ \cline{3-4}
& & Window size & 5, 7, 10 \\ \cline{3-4}
& & Number of epochs & 25, 50, 100 
 \\\hline\hline

\end{tabular}
\end{table*}

Feature extraction was used to identify and extract the most significant and informative patterns from data. The goal was to transform high-dimensional data into low-dimensional embeddings, which could then be fed as input into other models. Because there was no reliable ground truth, we relied solely on unsupervised feature extraction techniques. They are described next for each data source independently. The best hyperparameter values for all three data sources (tags, images, and diagnoses) were chosen based on clustering results obtained on the validation dataset. The starting values for each of the hyperparameter spans (where applicable) were selected based on the best practice found in various relevant literature (i.e.~papers proposing utilised methods).
The clustering process will be described in greater detail in section~\ref{ssec:experimental_setup}, while the following subsections discuss each data source's feature extraction.

\subsubsection{DICOM tags}\label{ssec:feature_extract_dcm_tag}
\newcommand{\tagsAElayers}{128 and 64}
\newcommand{\tagsBottleneckSize}{32}
\newcommand{\tagsMaxNrOfEpochs}{100}
\newcommand{\tagsEarlyStoppingPatience}{5}
\newcommand{\tagsBatchSize}{32}

DICOM tag feature extraction was done using principal component analysis (PCA)~\cite{abdi2010pca} and autoencoders (AEs)~\cite{napravnik2022autoencoders}. An extensive grid search of hyperparameters was performed for each approach. Multiple AEs having differing learning rates, architectures and bottleneck layer sizes were trained. All AEs consisted of three dense layers of differing sizes, each followed by a rectified linear unit (ReLU) activation function~\cite{agarap2018relu}. Across all trained AEs, the maximal number of epochs was {\tagsMaxNrOfEpochs} with mini-batch size {\tagsBatchSize}, and the chosen loss function was the mean squared error (MSE). If model training showed no loss improvement in {\tagsEarlyStoppingPatience} consecutive epochs, it was stopped.

Hyperparameter value ranges used in our experiments are shown in Table~\ref{tab:table_of_all_tested_hyperparams_for_all_sources}. Initially, when training AEs, all of the learning rates were tested. However, upon a more detailed inspection of the first hundred models, only learning rates $10^{-2}$ and $10^{-3}$ were used due to their superior performance. Moreover, only architectures with gradually decreasing layer sizes were tested.

\subsubsection{Images}\label{ssec:feature_extract_images}
\newcommand{\imageBatchSize}{$32$}
\newcommand{\imageEpochValidationCnt}{two}
\newcommand{\imageEarlyStoppingPatience}{5}
\newcommand{\imageNrEpochs}{40}

To obtain features from images, multiple neural network architectures commonly used in medical image processing~\cite{tajbakhsh2020med_image_seg_methods} were used. The tested modelling architectures were: convolutional autoencoder (CAE), the original U-Net~\cite{ronneberger2015unet}, recurrent residual convolutional neural network based on U-Net (R2U-Net)~\cite{alom2018r2unet} and U-Net coupled with the attention mechanism (AttU-Net)~\cite{oktay2018attentionunet}.

The training set was divided into mini-batches of size {\imageBatchSize} and, due to a large number of images, validation was performed {\imageEpochValidationCnt} times during a single epoch. Adam was used as the optimiser across all models, and the chosen loss function was MSE. All models were allowed to train for {\imageNrEpochs} epochs but were stopped if validation loss was not reduced in {\imageEarlyStoppingPatience} epochs. To see if any further dimensionality reduction could improve clustering results, PCA was applied with an extensive grid search of hyperparameters, as shown in Table~\ref{tab:table_of_all_tested_hyperparams_for_all_sources}.

U-Net, AttU-Net and R2U-Net were implemented as described in the original papers~\cite{ronneberger2015unet, alom2018r2unet, oktay2018attentionunet}, while the implemented CAE architecture closely follows a similar pattern to the U-Net's encoder layout.
The CAE encoder was comprised of four convolutional layers of $3 \times 3$ kernel size, followed by a ReLU activation function and a $2 \times 2$ max-pooling layer. The layers consisted of $64, 128, 256$ and $512$ filters, respectively. The final layer in the encoder was a convolutional layer of $1,024$ filters. This was followed by the decoder, which mirrored the encoder's layout. In all tested models, image features were flattened before performing clustering.

\subsubsection{Diagnoses}\label{ssec:feature_extract_diagnoses}

To extract feature vectors from narrative diagnoses texts, we tested the following methods: bag of words (BOW) ~\cite{Mikolov:2013:cbow}, term frequency-inverse document frequency (TF-IDF)~\cite{Dang:2020:TDIDF}, and doc2vec~\cite{Mikolov:2014:Doc2Vec}. The utilised methods were selected based on surveys on popular methods for creating embeddings from text (and specifically clinical diagnoses)~\cite{KHATTAK:2019:survey_text, Kowsar:2019:survey_text2, Li:2022:survey_text_3}. We have also put a constraint on methods' complexity, dataset availability, and computational/time demands. Following these constraints, experimenting with big neural networks such as BERT or GPTs~\cite{DBLP:2018:BERT, DBLP:2020:GPT} became unfeasible and was left for future research. 

Hyperparameter value ranges used for processing diagnoses are provided in Table~\ref{tab:table_of_all_tested_hyperparams_for_all_sources}. Each method requires a word corpus, i.e.~a list of words used to build embeddings by the selected methods. The word corpus was built based on the words presented in the training dataset. However, for a word to be selected for the corpus, the number of its occurrences had to be greater or equal to the experimentally established threshold \textit{minimum word frequency}.

Each of the selected methods has its pros and cons; for instance, BOW is the simplest, fastest and computationally least demanding; but it treats all words equally no matter how many times they occur in the diagnosis. TF-IDF solves this issue by considering word frequencies but still lacks context representation among the words. Finally, doc2vec methods are based on training a shallow neural network to predict: (1) the following word, using \textit{paragraph vectors: distributed memory} (PV-DM); or, (2) words belonging to the given diagnosis, using \textit{distributed BOW version of paragraph vector} (PV-DBOW). Hence, they can learn the content of a given paragraph by learning the connection between the words present in the paragraph. The input for PV-DM is the paragraph vector (embedding) and context words selected by the window size hyperparameter, while for the PV-DBOW input is only the paragraph vector. Length of learned paragraph vectors (embeddings) are defined by hyperparameter embedding size.

\subsection{Experimental setup}\label{ssec:experimental_setup}
\newcommand{\maxNrClusterIterations}{300}
\newcommand{\clusterCounts}{5, 10, 15, 20, 25, 30, 40, 50, 75, 100, 150}

Clustering was performed separately on all three sources of data (tags, images, and diagnoses). First, raw data were preprocessed and fed into the described feature extractors, after which the extracted feature embeddings were clustered.

Two clustering algorithms were used: k-means~\cite{elkan2003kmeans} and k-medoids~\cite{rdusseeun1987clusteringMedoids}.  For k-medoids, two different distance metrics were used: cosine distance and Euclidean distance. Clustering was performed for $\kappa \in \{\clusterCounts\}$ number of clusters. We also experimented with larger values of $\kappa$ but eventually left them out of the experimental setup due to some issues, such as obtaining a large number of empty clusters, significant overlapping between data points from different clusters, or other indications of overfitting.

\subsubsection{Evaluation metrics} \label{ssec:metrics_for_evaluation}

We assumed that optimal clustering results would include perfect homogeneity regarding the imaging modality and the examined body part. Hence, to measure the effectiveness of clustering, homogeneity score (HS) and normalised mutual information (NMI) were calculated for \textit{Modality} and \textit{BodyPartExamined} tags. Both of these metrics have a range of $[0.00, 1.00]$, where $0$ is the lowest and $1$ is the highest score. 
If we denote $y_M$ as the imaging modality, $\hat{y}$ as the predicted cluster label, $I(y_M, \hat{y})$ as mutual information between the two, $H(y_M)$ and $H(\hat{y})$ as their entropy, then NMI regarding modality ($NMI_M$) can be calculated as \cite{Kvaalseth2017nmi}:
\begin{equation}
    NMI_M = \frac{2 \cdot I(y_M,\hat{y})}{H(y_M) + H(\hat{y})}.
\end{equation}
$I(y_M, \hat{y})$ can be calculated as $I(y_M, \hat{y}) = H(y_M) - H(y_M | \hat{y})$, where $H(y_M|\hat{y})$ is the conditional entropy. HS regarding modality ($HS_M$) can be calculated as:
\begin{equation}
    HS_M = 1 - \frac{H(y_M|\hat{y})}{H(y_M)}.
\end{equation}

\noindent It is important to note that the denominator in the equation for calculating $HS_M$ can never be $0$ because the observed subset is not perfectly balanced (the observed "subset is not monotonically pure"). The same process applies when calculating NMI and HS regarding the examined body part ($NMI_B$ and $HS_B$, respectively), with the exception that $y_M$ is replaced by $y_B$, the \textit{BodyPartExamined} tag. The predicted cluster label $\hat{y}$ is always the same.

Finally, overall clustering quality was assessed by calculating the harmonic mean of all four metrics: $HS_B$, $HS_M$, $NMI_B$, and $NMI_M$. This harmonic mean, henceforth referred to as score $S$, provided a comprehensive evaluation of the grouping quality by considering all four metrics simultaneously.

Other than being homogeneous regarding imaging modality and examined body part, optimal clustering results should also exhibit similarities between images and diagnoses. We would expect that all data points within the same cluster display a visible similarity when comparing images and also show that their respective diagnoses carry related information and contain similar wording. To test this, cosine distances for image and diagnoses embeddings were calculated. To calculate similarities between images, the following was performed. First, let us assume there are $k$ data points assigned to cluster with index $c$, $0 \leq c < \kappa$. For each pair $(i, j), i \neq j$ from cluster $c$, whose images are denoted as $x_I^{(i)}$ and $x_I^{(j)}$ respectively, and their embeddings as $f(x_I^{(i)})$ and $f(x_I^{(j)})$, find the cosine distance as:

\begin{equation}
    d(x_I^{(i)}, x_I^{(j)}) = 1 - \frac{ f(x_I^{(i)})^T \cdot f(x_I^{(j)}) }{ \| f(x_I^{(i)}) \| \cdot \| f(x_I^{(j)}) \| }.
\end{equation}

The possible number of pairs in cluster $c$ is $u^{(c)}=\frac{k}{2}(k-1)$. To find the dissimilarity of images in cluster $D_I^{(c)}$, calculate the mean cosine distance of all pairs from cluster $c$:
\begin{equation}
    D_I^{(c)} = \frac{1}{u^{(c)}} \sum_{i = 0}^{k-1} \sum_{j = i+1}^{k-1} d(x_I^{(i)}, x_I^{(j)}),
\end{equation}
and finally, overall image similarities across all clusters were calculated as $D_I = \frac{1}{\kappa} \sum_{c = 0}^{\kappa-1} D_I^{(c)}$.

The same process was applied to get diagnoses similarities $D_{D}$ from diagnoses embeddings. Ideally, $D_{I}$ and $D_{D}$ should be close to $0$, or in other words, the distances between embeddings in the same cluster should be as small as possible.

\subsubsection{Evaluation process}

\begin{figure}[!tb]
    \centering
    \includegraphics[width=\linewidth]{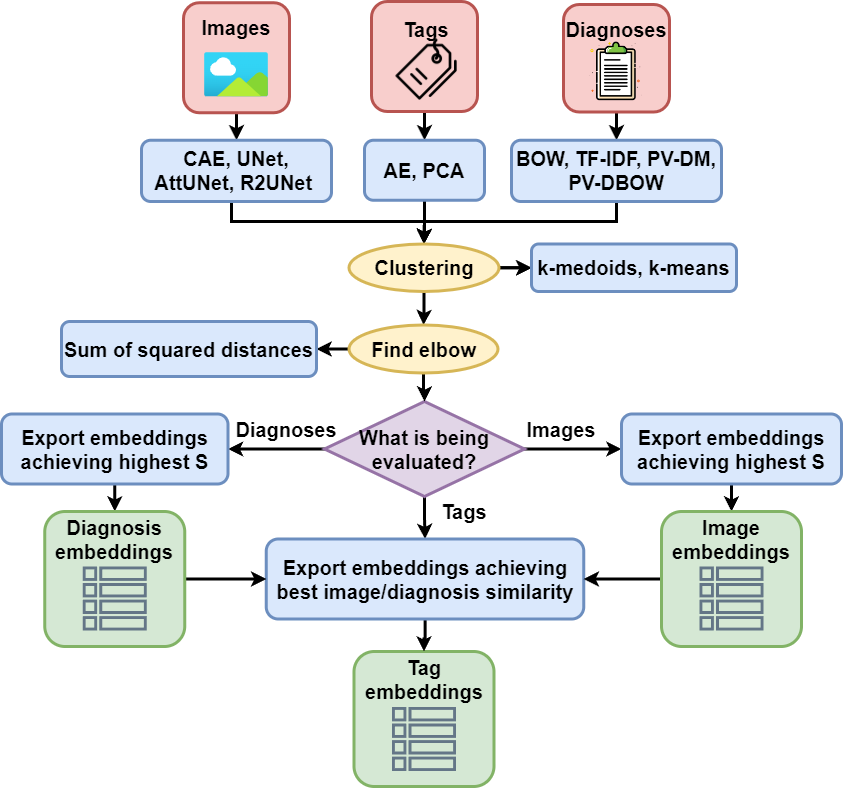}
    \caption{Embedding evaluation pipeline for all three data sources. Diagnoses and images were evaluated by their homogeneity and mutual information of modality and examined body part, which are both tags found in DICOM metadata. On the other hand, DICOM tags were evaluated on the (dis)similarity of diagnoses and images in the obtained groups. Namely, due to the nature of the DICOM standard \cite{dicom_standard} and the frequent occurrence of modality-specific values, it was deemed to be more objective if the DICOM tags were evaluated on diagnoses and image embedding (dis)similarity. In short, the images and diagnoses were clustered separately. The best-performing embeddings were chosen and then used to evaluate the performance of DICOM tag clustering. This was done to make a more objective assessment of different data sources' embeddings. }
    \label{fig:evaluation_pipeline_of_data_sources}
\end{figure}

To find the best individual data source embeddings, we compared clustering performance on the validation set across all data sources and all feature extractors. The overall process is illustrated in Fig.~\ref{fig:evaluation_pipeline_of_data_sources}. Initially, to find the optimal number of clusters for each of the data sources, the elbow method \cite{thorndike1953elbow_belongs} was utilised on the sum of squared distances of data points to their closest cluster centre. To detect the elbow, we used the Kneedle algorithm by Satopaa et al.~\cite{satopaa2011finding_knee}. Having too few clusters could result in heterogeneous grouping, while having too many clusters might lead to groups that are homogeneous but show evidence of incompleteness~\cite{rosenberg2007v}.

As was described in section~\ref{ssec:metrics_for_evaluation}, the metrics used for clustering evaluation were: modality homogeneity ($HS_M$), body part examined homogeneity ($HS_B$), modality normalised mutual information ($NMI_M$), body part examined normalised mutual information ($NMI_B$), image embedding similarity ($D_I$) and diagnosis embedding similarity ($D_D$). To adequately analyse the clustering results, different sources of data were evaluated using different metrics. The efficiency of DICOM tag clustering was evaluated on image and diagnosis similarities. On the other hand, image and diagnosis clustering was evaluated on how homogeneous the results were regarding the imaging modality and body part examined, namely, using the metrics $NMI_B$, $NMI_M$, $HS_B$ and $HS_M$.

Following image embedding clustering and finding the optimal number of clusters, $NMI_B$, $NMI_M$, $HS_B$ and $HS_M$ were compared. The feature extractor achieving the highest $S$ score at the elbow was chosen for later analysis.  The exact same rules were applied simultaneously to diagnosis embeddings, where the best achieving model was chosen based on their $S$ at the elbow. 

Following the same pattern as image and diagnosis grouping, DICOM tags were clustered, and then the elbow method was applied. After this, visual similarities $D_{I}$ and textual similarities $D_{D}$ were calculated at the elbow. For this purpose, the best-performing image and diagnosis embeddings from the previous step were used. Finally, to rank the efficiency of DICOM tag feature extraction models, we used the metric $D_{score}$ as the harmonic mean of $D_I$ and $D_D$. 
The primary objective is to create clusters that exhibit the highest degree of data similarity. Consequently, it is necessary to quantify and recognise the similarity among data instances within each identified cluster. Therefore, the model obtaining the lowest $D_{score}$ value at the elbow would be selected as the best DICOM tag feature extraction model.

\subsubsection{Feature fusion}
After selecting the best feature extractor for each of the data sources, the resulting embeddings were combined in three ways: direct concatenation, concatenation of cluster-space distances, and concatenation of cluster probability assignments. 
In each of the approaches, the resulting vector of a single data point $i$ was flat, and in the format of $f(x^{(i)})=\Big[f(x^{(i)}_D),f(x^{(i)}_T),f(x^{(i)}_I)\Big]^T$, where $f(x^{(i)}_D)$ is the diagnosis embedding, $f(x^{(i)}_T)$ is the DICOM tags embedding, and  $f(x^{(i)}_I)$ is the image embedding.

The first and simplest approach was to concatenate the raw embeddings. Embeddings from each of the data sources were concatenated into a single, flat vector. 
The second approach was to use cluster-space distances. When clustering a single embedding, distances to each of the cluster centres are computed, and then the point is assigned to the nearest cluster. Embeddings carrying similar information should also have similar distances to each of the cluster centres. Hence, instead of using the extracted embeddings, we computed the distances to each of the cluster centres and then concatenated those distances together to fuse the data sources. All distances were normalised to fit the range $[0.00, 1.00]$ before concatenation. For ease of reference, this approach will henceforth be referred to as \textit{clusterdists}. 
The third approach was closely related to the previous one, albeit with a small difference. Instead of using the distances as they are, we computed their probability assignments for each cluster. The probability of $i$-th instance being assigned to cluster $k$ is calculated using the softmax function, assigning higher probabilities to shorter distances:

\begin{equation}
    p_k^{(i)} = \frac{e^{-d_k^{(i)}}}{\sum_{j=1}^{\kappa}e^{-d_j^{(i)}}},
\end{equation}
where $\kappa$ is the number of clusters, and $d_k$ and $d_j$ are distances to $k$-th and $j$-th cluster of the respective source embeddings, respectively.
For ease of reference, this approach will henceforth be referred to as \textit{clusterprobs}.

Other methods of feature fusion, such as the approach used by Radford et al.~\cite{radford2021clip}, were considered as potential options. However, these approaches were eventually rejected due to constraints posed by hardware limitations and the volume of data involved.

\section{Results on experimental subset}\label{sec:results}

\begin{table*}[!tbh]
\centering
\footnotesize
\caption{Hyperparameter values of best performing feature extraction models (emphasised) for each data source independently and all feature extractors for respective sources covered by our experiments (Table~\ref{tab:table_of_all_tested_hyperparams_for_all_sources}).}
\label{tab:best_models_hyperparameters_all_data_sources}

\begin{tabular}{cclcccc}
\hline\hline
\textbf{Data source} & \textbf{Model name} & \textbf{Hyperparameters} & \textbf{\begin{tabular}[c]{@{}c@{}}Embedding\\ size\end{tabular}} & \textbf{Algorithm} & \textbf{\begin{tabular}[c]{@{}c@{}}Cluster\\ metric\end{tabular}} & \textbf{\begin{tabular}[c]{@{}c@{}}Number of\\ clusters\end{tabular} } \\ \hline


\multirow{8}*{Images} & \textbf{CAE} & Learning rate: $10^{-6}$ & $500$ & k-means & Euclidean & $40$ \\
 &  & PCA solver: randomised &  &  &  &  \\ 
 & U-Net & Learning rate: $10^{-6}$ & $10,000$ & k-means & Euclidean & $30$ \\
 &  & PCA solver: randomised &  &  &  &  \\ 
 & AttU-Net & Learning rate: $10^{-6}$ & $65,536$ & k-medoids & cosine & $30$ \\
 &  & PCA not applied &  &  &  &  \\ 
 & R2U-Net & Learning rate: $10^{-6}$ & $5,000$ & k-medoids & cosine & $40$ \\
 &  & PCA solver: randomised &  &  &  & \\
 \hline 

\multirow{8}{*}{Diagnoses}& TF-IDF & Minimum word frequency: $1,000$ & $825$ & k-means & Euclidean & $30$ \\
& BOW & Minimum word frequency: $10$ & $7,951$ & k-means & Euclidean & $30$ \\
& PV-DM & Minimum word frequency: $100$ & $10$ & k-means & Euclidean & $20$ \\
&  & Window size: $5$ &  &  &  &  \\
&  & Number of epochs: $50$ &  &  &  &  \\
& \textbf{PV-DBOW} & Minimum word frequency: $50$ & $1,000$ & k-means & Euclidean & $20$ \\
&  & Window size: $7$ &  &  &  &  \\
&  & Number of epochs: $50$ &  &  &  & \\ 
\hline

\multirow{3}{*}{DICOM tags} & \textbf{AE} & $512 \rightarrow 200 \rightarrow 125 \rightarrow 32$ & 32 & k-medoids & cosine & 25 \\
& & Learning rate: $10^{-2}$ &  &  &  &  \\
& PCA & Solver: ARPACK & 50 & k-medoids & Euclidean & 40

\\ \hline\hline
\end{tabular}
\end{table*}

\begin{table*}[!tb]
\centering
\footnotesize
\caption{Results for each best performing feature extraction model for images, diagnoses and tag clustering, computed on the validation dataset. Best results are emphasised. $\pm$ sign delimits the mean from the standard deviation.}
\label{tab:individual_best_models_clustering_results}
\begin{tabular}{cccccccccc}
\hline\hline
\textbf{Data source} & \textbf{\begin{tabular}[c]{@{}c@{}}Model\\ name\end{tabular}} & \textbf{$NMI_B$} & \textbf{$NMI_M$} & \textbf{$HS_B$} & \textbf{$HS_M$} & \textbf{$S$} & \textbf{$D_{D} [\cdot 10^{-2}]$} & \textbf{$D_{I} [\cdot 10^{-2}]$} & \textbf{$D_{score}$} \\

\hline

\multirow{4}{*}{Images} & \textbf{CAE} & \textbf{0.430} & \textbf{0.459} & \textbf{0.538} & \textbf{0.744} & \textbf{0.519} & - & - & - \\
& U-Net & 0.401 & 0.450 & 0.467 & 0.673 & 0.479 & - & - & - \\
& AttU-Net & 0.285 & 0.348 & 0.328 & 0.512 & 0.351 & - & - & - \\
& R2U-Net & 0.186 & 0.206 & 0.185 & 0.255 & 0.205 & - & - & - \\

\hline 
\multirow{4}{*}{Diagnoses} & TF-IDF & \textbf{0.603} & 0.493 & 0.623 & 0.716 & 0.598 & - & - & - \\
& BOW & 0.529 & 0.508 & 0.576 & 0.788 & 0.583 & - & - & - \\
& PV-DM & 0.544 & 0.467 & 0.570 & 0.689 & 0.557 & - & - & - \\
& \textbf{PV-DBOW} & 0.592 & \textbf{0.569} & \textbf{0.634} & \textbf{0.77} & \textbf{0.633} & - & - & - \\

\hline

\multirow{2}{*}{DICOM tags} & \textbf{AE} & - & - & - & - & - &  \textbf{3.553 $\pm$ 2.11} &  \textbf{3.051 $\pm$ 1.01} &  \textbf{3.283} \\
& PCA & - & - & - & - & - & 3.578 $\pm$ 2.019 & 3.248 $\pm$ 1.171  &  3.405

\\ \hline\hline
\end{tabular}
\end{table*}

Next, we present the results concerning the evaluation of embeddings clustering generated for each of the three data sources. First, we present the results for individual sources in section~\ref{sec:optimal_embeddings}. Next, in section~\ref{sec:AblationStudy}, the ablation study concerning fusing individual source embeddings is given. Lastly, we discuss our findings concerning the experiments in section~\ref{sec:dis}.

\subsection{Optimal embeddings} \label{sec:optimal_embeddings}

\begin{figure*}[!tb]
    \centering
    \includegraphics[width=\linewidth]{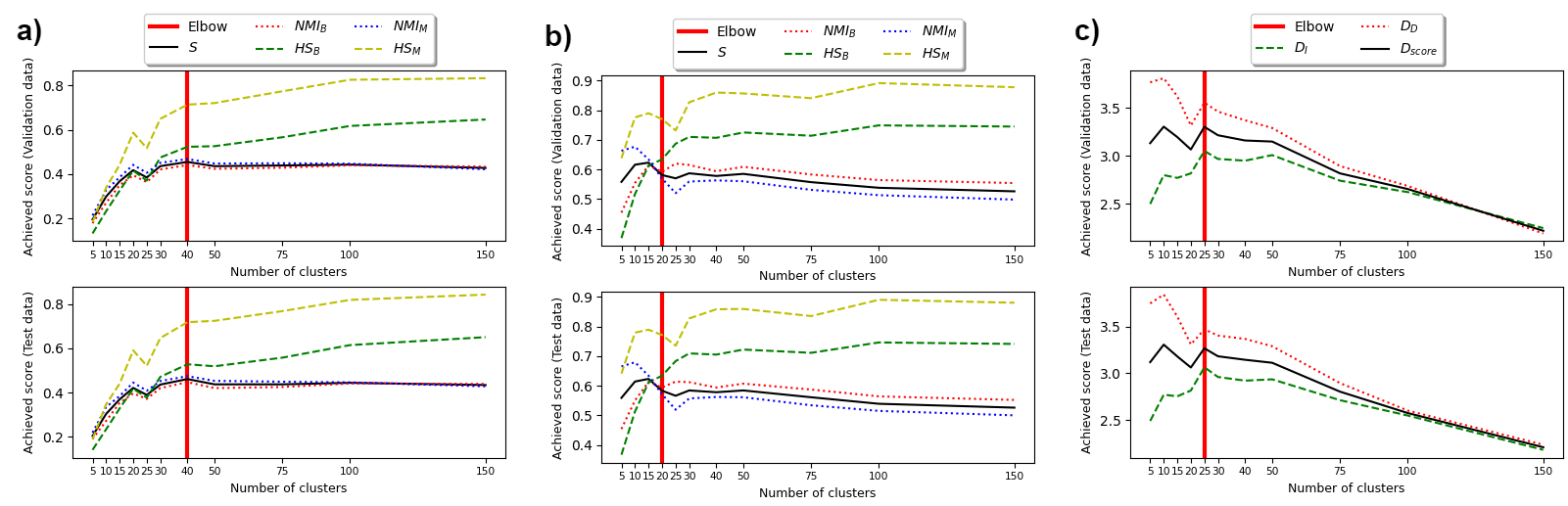}
    \caption{Diagrams showing individual evaluation metrics values on validation (top) and test (bottom) subsets, when clustering optimal image embeddings (CAE, subfigure a)), optimal diagnoses embeddings (PV-DBOW, subfigure b)) and DICOM tag embeddings (AE, subfigure c)). For all the data sources and extractors, the performance is almost identical on both datasets.}
    \label{fig:all_individual_metrics_all_nclusters}
\end{figure*}

As was described in section~\ref{ssec:feature_extraction},
we have trained four different types of models for clustering textual diagnoses (TF-IDF, BOW, PV-DM, PV-DBOW), accompanied by four different model architectures for image feature extraction (CAE, U-Net, AttU-Net, R2U-Net) and two different extractors for DICOM tags (AE, PCA). Embeddings obtained by each of these extractors were tested on all hyperparameter values and across all clustering setups.
In Table~\ref{tab:best_models_hyperparameters_all_data_sources}, we present the best hyperparameter values for each of the four tested models types, while in Table~\ref{tab:individual_best_models_clustering_results}, we provide the results for the selected best-performing models.

Among the image extractors, results single out CAE as the best-performing model. CAE outperformed U-Net, AttU-Net and R2U-Net in terms of modality and examined body part homogeneity. It obtained the highest $HS_M$, $HS_B$, $NMI_M$ and $NMI_B$, and thus the highest $S$ score, on the validation dataset.

Regarding extractors for narrative diagnoses, PV-DBOW attained the highest $HS_M$, $HS_B$ and $NMI_M$ scores. While its $NMI_B$ was second only to TF-IDF, the overall $S$ score shows that PV-DBOW outperformed the other models.

To calculate image and diagnoses distances (and the corresponding $D_{score}$) for DICOM tag evaluation, the best-performing feature extractors from images and diagnoses were used. In other words, CAE embeddings were used to calculate image similarities, while PV-DBOW embeddings were utilised for diagnoses comparison.
As it can be seen in Table~\ref{tab:individual_best_models_clustering_results}, the best image and diagnosis similarity on the validation dataset was achieved using AE.

A more detailed performance of best-performing models' clustering results is shown in Fig.~\ref{fig:all_individual_metrics_all_nclusters}, where Fig.~\ref{fig:all_individual_metrics_all_nclusters}a) shows CAE performance as the highest scoring image feature extractor, Fig.~\ref{fig:all_individual_metrics_all_nclusters}b) shows the same for diagnoses (PV-DBOW) and Fig.~\ref{fig:all_individual_metrics_all_nclusters}c) for DICOM tags (AE). From the shown metrics, it is evident that the models perform nearly the same on the validation and test sets, showcasing that the models did not overfit.

\begin{table*}[!tb]
\centering
\caption{Hyperparameter values of each best performing model obtained by fusing individual source embeddings. Model performance is shown in Table~\ref{tab:results_on_all_data_sources_and_fused_features_val_data}).
In this table,\textit{AE} is used to describe DICOM tag embeddings, \textit{CAE} as image embeddings as \textit{PV-DBOW} as diagnosis embeddings.}
\label{tab:params_of_best_models_for_fused_features_clustering}
\footnotesize
\begin{tabular}{cccccc}
\hline\hline
\textbf{Model name} & \textbf{Combine Method} & \textbf{Embedding size} & \textbf{Algorithm} & \textbf{Clustering metric} & \textbf{Num clusters} \\ \hline

\multirow{3}{*}{{[}AE{]}-{[}CAE{]}} & clusterdists & 65 & k-means & Euclidean & 30 \\
 & clusterprobs & 65 & k-medoids & cosine & 30 \\
 & embeddings & 532 & k-means & Euclidean & 75 \\
\hline

\multirow{3}{*}{{[}PV-DBOW{]}-{[}AE{]}} & clusterdists & 45 & k-means & Euclidean & 30 \\
 & clusterprobs & 45 & k-medoids & Euclidean & 30 \\
 & embeddings & 1032 & k-means & Euclidean & 40 \\
 \hline
 
\multirow{3}{*}{{[}PV-DBOW{]}-{[}CAE{]}} & clusterdists & 60 & k-medoids & cosine & 30 \\
 & clusterprobs & 60 & k-medoids & Euclidean & 40 \\
 & embeddings & 1500 & k-means & Euclidean & 40 \\
 \hline
 
\multirow{3}{*}{{[}PV-DBOW{]}-{[}AE{]}-{[}CAE{]}} & clusterdists & 85 & k-medoids & Euclidean & 40 \\
 & clusterprobs & 85 & k-medoids & cosine & 40 \\
 & embeddings & 1532 & k-means & Euclidean & 50 

 \\ \hline\hline
\end{tabular}

\end{table*}

\begin{table*}[!tb]
\footnotesize
\centering
\caption{Results for each best-performing model for all combinations of data sources, computed on the validation dataset. Best results are emphasised for each specific metric utilised. $\pm$ sign delimits the mean from the standard deviation. In this table, \textit{AE} is used to describe DICOM tag embeddings, \textit{CAE} as image embeddings as \textit{PV-DBOW} as diagnosis embeddings.}
\label{tab:results_on_all_data_sources_and_fused_features_val_data}
\begin{tabular}{cccccccccc}
\hline\hline
\textbf{Model name} & \textbf{Combine Method} & \textbf{$NMI_B$} & \textbf{$NMI_M$} & \textbf{$HS_B$} & \textbf{$HS_M$} & \textbf{$S$} & \textbf{$D_{D} [\cdot 10^{-2}]$} & \textbf{$D_{I} [\cdot 10^{-2}]$} & \textbf{$D_{score}$} \\ \hline

\multirow{3}{*}{{[}AE{]}-{[}CAE{]}} & clusterdists & 0.486 & 0.652 & 0.574 & 0.99 & 0.631 & 3.599 $\pm$ 1.976 & 2.663 $\pm$ 1.129 & 3.061 \\
 & clusterprobs & 0.503 & 0.68 & 0.578 & 0.999 & 0.647 & 3.564 $\pm$ 2.061 & 2.759 $\pm$ 1.101 & 3.11 \\
 & embeddings & 0.462 & 0.534 & 0.613 & 0.928 & 0.593 & 3.232 $\pm$ 1.787 & 2.268 $\pm$ 0.976 & 2.666 \\
 \hline
 
\multirow{3}{*}{{[}PV-DBOW{]}-{[}AE{]}} & clusterdists & 0.46 & 0.675 & 0.532 & 0.999 & 0.612 & \textbf{2.894 $\pm$ 2.356} & 2.717 $\pm$ 1.039 & 2.802 \\
 & clusterprobs & 0.516 & \textbf{0.697} & 0.581 & \textbf{1.0} & 0.656 & 3.385 $\pm$ 2.205 & 2.802 $\pm$ 1.045 & 3.066 \\
 & embeddings & 0.609 & 0.563 & 0.712 & 0.845 & \textbf{0.666} & 3.228 $\pm$ 1.891 & 2.179 $\pm$ 0.863 & 2.602 \\

 \hline
\multirow{3}{*}{{[}PV-DBOW{]}-{[}CAE{]}} & clusterdists & 0.373 & 0.429 & 0.44 & 0.651 & 0.453 & 2.927 $\pm$ 2.038 & 2.457 $\pm$ 0.965 & 2.671 \\
 & clusterprobs & 0.425 & 0.453 & 0.529 & 0.732 & 0.512 & 3.269 $\pm$ 1.878 & 2.272 $\pm$ 1.031 & 2.681 \\
 & embeddings & \textbf{0.611} & 0.548 & \textbf{0.713} & 0.819 & 0.657 & 3.234 $\pm$ 1.957 & 2.163 $\pm$ 0.879 & 2.592 \\

 \hline
\multirow{3}{*}{\textbf{{[}PV-DBOW{]}-{[}AE{]}-{[}CAE{]}}} & clusterdists & 0.472 & 0.616 & 0.583 & 0.986 & 0.618 & 3.362 $\pm$ 2.088 & 2.587 $\pm$ 1.104 & 2.924 \\
 & clusterprobs & 0.497 & 0.639 & 0.604 & \textbf{1.0} & 0.642 & 3.483 $\pm$ 1.996 & 2.779 $\pm$ 1.134 & 3.091 \\
 & \textbf{embeddings} & 0.587 & 0.57 & 0.707 & 0.885 & \textbf{0.666} & 2.982 $\pm$ 1.688 & \textbf{2.012 $\pm$ 0.783} & \textbf{2.403}
 \\ \hline\hline
\end{tabular}
\end{table*}

\begin{figure}[!tb]
    \centering
    \includegraphics[width=0.9\linewidth]{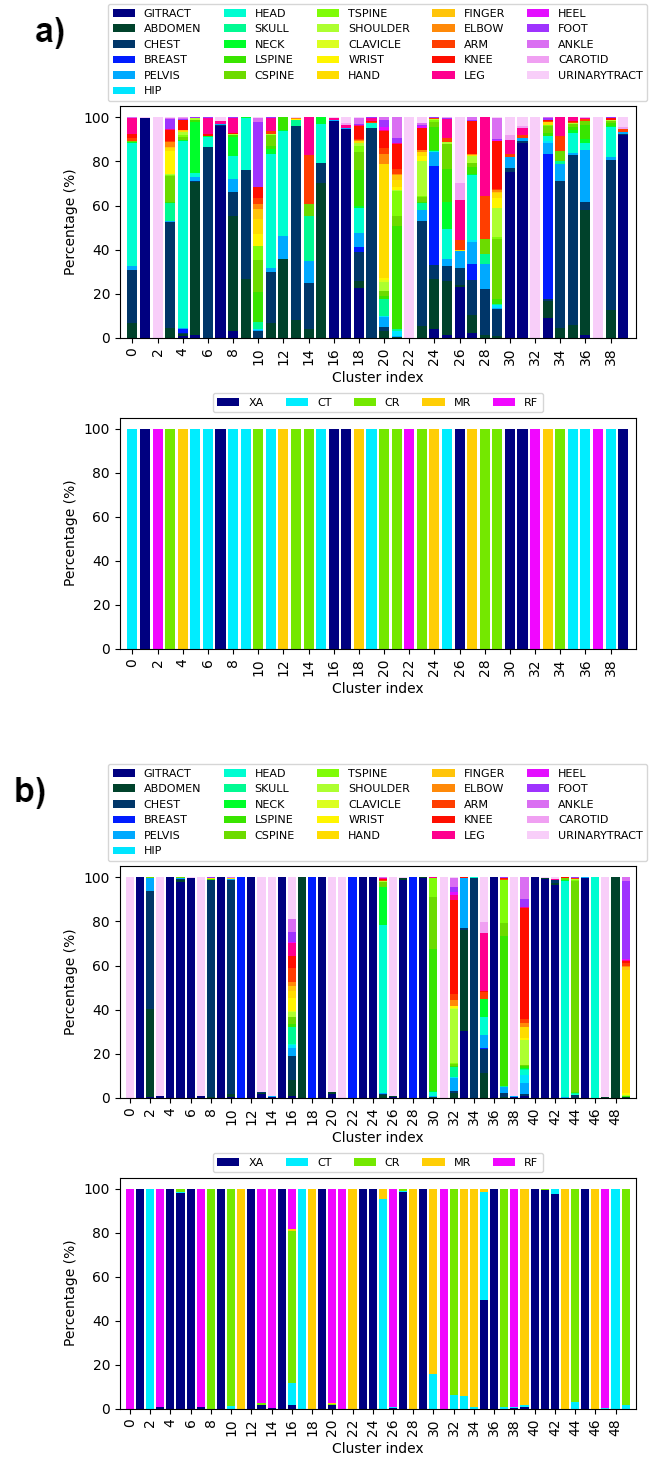}
    \caption{Grouping quality regarding modality and examined body part, when grouping by [PV-DBOW]-[AE]-[CAE] using \textit{clusterprobs} (subfigure a)) and \textit{embeddings} (subfigure b)) combine methods. In these plots, each bar represents the mixture ratio within a specific cluster. In subfigures a) and b), the first image shows how homogeneous the clusters are when observing the body part (i.e.~how mixed the clusters are with regard to anatomical region), while the second one shows the different modalities in each cluster (i.e.~how mixed the clusters are with regard to imaging modality).}
    \label{fig:dataset_grouping_quality_by_mod_and_bpe_pvdbow_ae_cae}
\end{figure}

\subsection{Source fusion ablation study}\label{sec:AblationStudy}

The chosen models for feature fusion were AE for DICOM tags, CAE for images and PV-DBOW for diagnoses.
We performed an extensive study to see how the clustering results change based on which features are included, and which are not. 
An analysis of hyperparameters was performed on the validation set, based on which the best hyperparameter values were chosen and are shown in  Table~\ref{tab:params_of_best_models_for_fused_features_clustering}, while their respective performance on the validation set can be seen in Table~\ref{tab:results_on_all_data_sources_and_fused_features_val_data}. 
The best hyperparameters were chosen primarily based on their performance regarding the metric $S$, while the metric $D_{score}$ was taken into consideration where the metric $S$ was deemed insufficient to adequately distinguish between the best results.

DICOM tags and images ([AE]-[CAE]): When observing results given in Table~\ref{tab:results_on_all_data_sources_and_fused_features_val_data} and those in Table~\ref{tab:individual_best_models_clustering_results}, it becomes apparent that grouping DICOM tags with images leads to an improvement in the $D_{score}$ compared to using AE alone. Moreover, all three combine methods (\textit{embeddings}, \textit{clusterdists} and \textit{clusterprobs}) yield a higher $S$ score than using images alone, exhibiting better modality and examined body part homogeneity.

Diagnoses and DICOM tags ([PV-DBOW]-[AE]): Following a similar pattern to [AE]-[CAE], combining DICOM tags with diagnoses leads to an improvement in the $D_{score}$ compared to using DICOM tags alone. When applying the \textit{embeddings} combination method, the $S$ score is higher than the one obtained using just diagnosis (and is the highest obtained overall on the validation dataset), with visible improvement, particularly in $HS_B$ and $HS_M$.
Moreover, when observing the \textit{clusterdists} and \textit{clusterprobs} methods, the combined approach exhibits a notable increase in modality homogeneity compared to just using diagnoses embeddings. However, there is a trade-off in terms of examined body part homogeneity, as $NMI_B$ and $HS_B$ are lower in the \textit{clusterdists} and \textit{clusterprobs} approaches than the $HS_B$ and $NMI_B$ obtained by diagnoses alone.

Diagnoses and images ([PV-DBOW]-[CAE]): Combining images with diagnoses, particularly using the \textit{embeddings} method, results in the best grouping by examined body part; in this manner, the overall highest $NMI_B$ and $HS_B$ are achieved. The $S$ score obtained through this combination is better than the $S$ score achieved by using images and diagnoses independently.

Diagnoses, DICOM tags and images ([PV-DBOW]-[AE]-[CAE]): Lastly, when all three data sources (images, DICOM tags, diagnoses) are combined using \textit{embeddings} method, the best overall $S$ score is achieved. This score is equal to the $S$ score obtained when using [PV-DBOW]-[AE] (diagnoses and DICOM tags) using the \textit{embeddings} approach; however there is a clear difference in $D_{score}$ between them.
On the other hand, when all three data sources are combined using the \textit{clusterprobs} method, a perfect score for $HS_M$ is obtained.
Fig.~\ref{fig:dataset_grouping_quality_by_mod_and_bpe_pvdbow_ae_cae}a) and Fig.~\ref{fig:dataset_grouping_quality_by_mod_and_bpe_pvdbow_ae_cae}b) show how different quality of groupings are obtained through these two different combine methods (\textit{clusterprobs} versus~\textit{embeddings}).

Finally, the performance of all individual data sources, as well as all feature combinations and combination methods on the test dataset is given in Table~\ref{tab:results_on_all_data_sources_and_fused_features_test_data}.

\begin{table*}[!tb]
 \centering
\caption{Clustering results on the test dataset, when using the best performing models from all data sources and all three feature fusion approaches. $\pm$ sign delimits the mean from the standard deviation.}
\label{tab:results_on_all_data_sources_and_fused_features_test_data}
\footnotesize
\begin{tabular}{cccccccccc}
\hline\hline
\textbf{Model name} & \textbf{Combine Method} & \textbf{$NMI_B$} & \textbf{$NMI_M$} & \textbf{$HS_B$} & \textbf{$HS_M$} & \textbf{$S$} & \textbf{$D_{D} [\cdot 10^{-2}]$} & \textbf{$D_{I} [\cdot 10^{-2}]$} & \textbf{$D_{score}$} \\
\hline

AE (DICOM Tags) & - & - & - & - & - & - & 3.471 $\pm$ 2.059 & 3.065 $\pm$ 1.107 & 3.255 \\
CAE (Images) & - & 0.447 & 0.474 & 0.528 & 0.719 & 0.524 & - & - & - \\
PV-DBOW (Diagnoses) & - & 0.594 & 0.571 & 0.634 & 0.771 & 0.634 & - & - & - \\
\hline

\multirow{3}{*}{{[}AE{]}-{[}CAE{]}} & clusterdists & 0.494 & 0.652 & 0.584 & 0.992 & 0.637 & 3.592 $\pm$ 1.941 & 2.654 $\pm$ 1.146 & 3.053 \\
 & clusterprobs & 0.507 & 0.678 & 0.584 & \textbf{1.0} & 0.65 & 3.546 $\pm$ 2.026 & 2.769 $\pm$ 1.109 & 3.109 \\
 & embeddings & 0.465 & 0.539 & 0.615 & 0.933 & 0.597 & 3.22 $\pm$ 1.711 & 2.256 $\pm$ 0.983 & 2.653 \\
 \hline
 
\multirow{3}{*}{{[}PV-DBOW{]}-{[}AE{]}} & clusterdists & 0.462 & 0.671 & 0.536 & 0.999 & 0.614 & \textbf{2.906 $\pm$ 2.358} & 2.697 $\pm$ 1.058 & 2.797 \\
 & clusterprobs & 0.523 & \textbf{0.69} & 0.593 & \textbf{1.0} & 0.662 & 3.364 $\pm$ 2.172 & 2.801 $\pm$ 1.09 & 3.057 \\
 & embeddings & 0.605 & 0.561 & 0.711 & 0.847 & 0.664 & 3.172 $\pm$ 1.815 & 2.179 $\pm$ 0.921 & 2.584 \\
 \hline
 
\multirow{3}{*}{{[}PV-DBOW{]}-{[}CAE{]}} & clusterdists & 0.368 & 0.424 & 0.434 & 0.642 & 0.447 & 2.913 $\pm$ 1.991 & 2.442 $\pm$ 0.967 & 2.657 \\
 & clusterprobs & 0.427 & 0.455 & 0.532 & 0.736 & 0.514 & 3.279 $\pm$ 1.85 & 2.269 $\pm$ 1.039 & 2.682 \\
 & embeddings & \textbf{0.611} & 0.546 & \textbf{0.712} & 0.816 & 0.656 & 3.208 $\pm$ 1.92 & 2.182 $\pm$ 0.93 & 2.598 \\
 \hline

\multirow{3}{*}{\textbf{{[}PV-DBOW{]}-{[}AE{]}-{[}CAE{]}}} & clusterdists & 0.478 & 0.617 & 0.591 & 0.989 & 0.623 & 3.354 $\pm$ 2.096 & 2.571 $\pm$ 1.1 & 2.911 \\
 & clusterprobs & 0.499 & 0.637 & 0.606 & \textbf{1.0} & 0.643 & 3.47 $\pm$ 1.94 & 2.787 $\pm$ 1.173 & 3.091 \\
 & \textbf{embeddings} & 0.587 & 0.571 & 0.705 & 0.885 & \textbf{0.666} & 3.024 $\pm$ 1.591 & \textbf{2.011 $\pm$ 0.828} & \textbf{2.416}

 \\ \hline\hline
\end{tabular}
\end{table*}

\section{Discussion}\label{sec:dis}
One of the main challenges we encountered in our research on how to approach data annotation involved identifying proper distinct classes, or annotation ontology. First, we explored the utility of applying LOINC/RSCNA Radiology Playbook~\cite{vreeman2018loinc} for guiding the annotation process. However, upon inspection of the data it turned out that the current clinical practice at CHC Rijeka differs significantly from what is proposed in the standard. Similar attempts were made to conjure our own alternative annotation ontology. However, this idea was also abandoned because of other challenges that were difficult to solve, such as region overlapping. As a result, we abandoned the idea of structurally constraining the annotation process this way, leaving it completely to the optimisation process.

As can be seen in Table \ref{tab:results_on_all_data_sources_and_fused_features_test_data}, of all the three data sources, image clustering (CAE) exhibited the worst performance regarding modality and examined body part homogeneity. Upon visual inspection of the obtained clusters, we noticed that, although images can be visually similar, they often showcase different body parts captured by the same modality; or vice versa; they show the same body part but are captured in different modalities. Different windowing parameters can greatly influence the image as well, an example of which can be seen in cluster 2 shown in Fig.~\ref{fig:cluster_snippets_mosaic}. Although all images show a part of the torso, there are significant differences in pixel intensity, which could lead to confounded grouping and a lower $NMI_B$ score. This suggests that images alone do not provide sufficient information for semantically good grouping. However, the results shown in Table~\ref{tab:results_on_all_data_sources_and_fused_features_test_data} indicate that integrating image data into the grouping process reduces the $D_{score}$ and improves the visual quality of the clusters. Thus, although images may not individually possess enough information for optimal semantic grouping, their inclusion contributes to enhanced cluster representation.

Diagnoses (PV-DBOW) showed excellent results regarding the quality of anatomical region grouping. This could be explained by the fact that diagnoses' wording has a significant focus on the anatomical region being examined by describing illnesses or injuries affecting a specific body part. When used in conjunction with other data sources -- as shown in Table~\ref{tab:results_on_all_data_sources_and_fused_features_test_data} -- it is evident that the quality of grouping by examined body part is significantly improved. Therefore, it can be inferred that the integration of diagnoses contributes to better anatomical region grouping.

Wherever DICOM tags (AE) were used, there was a noticeable improvement with regard to modality homogeneity (Table~\ref{tab:results_on_all_data_sources_and_fused_features_test_data}). When coupled with DICOM tags, images achieved  better $NMI_M$ and $HS_M$ results than they did independently. This is visible with diagnoses as well, where $HS_M$ increases when they are joined with DICOM tags. It is evident that DICOM tags contribute to better modality homogeneity, which can be explained by analysing the DICOM standard~\cite{dicom_standard} and observing that DICOM tags often have modality-specific values.

Three different feature fusion approaches were tested, all three showing satisfactory performance. Two alternative methods of feature fusion, namely \textit{clusterprobs} and \textit{clusterdists}, were introduced and tested. To the best of our knowledge, no similar feature fusion techniques have been used before. When compared to the \textit{embeddings} approach, \textit{clusterdists} and \textit{clusterprobs} favoured high modality homogeneity. In particular, three different models shown in Table~\ref{tab:results_on_all_data_sources_and_fused_features_test_data} ([PV-DBOW]-[AE]-[CAE] \textit{clusterprobs}, [AE]-[CAE] \textit{clusterprobs} and [PV-DBOW]-[AE] \textit{clusterprobs}) achieved perfect scores for $HS_M$. Nonetheless, the \textit{embeddings} approach consistently outperformed the others in terms of anatomical region homogeneity. This is especially visible in Fig.~\ref{fig:dataset_grouping_quality_by_mod_and_bpe_pvdbow_ae_cae}a) and~\ref{fig:dataset_grouping_quality_by_mod_and_bpe_pvdbow_ae_cae}b), where the former shows how it prioritises modality homogeneity, while the latter exhibits superior grouping in terms of the anatomical region. 

As is visible in Table~\ref{tab:results_on_all_data_sources_and_fused_features_test_data}, some models achieved similar $S$ scores. However, between these models there is a clear $D_{score}$ difference, with lower $D_{score}$ meaning that clusters are more visually homogeneous and contain more similar diagnoses, which is favourable. 
Thus, the approach utilising all three data sources combined using the \textit{embeddings} method can be considered the most efficient for achieving the best grouping results in terms of modality, examined body part, and image/diagnosis similarity.

\section{RadiologyNET dataset}\label{ssec:radiologynet_data_set}
\newcommand{\radiologynetSize}{1,337,926}
\newcommand{\radiologynetLargestClusterSize}{341,083}
\newcommand{\radiologynetSmallestClusterSize}{6}
\newcommand{\radiologynetHeterogeneousClusters}{16, 32, 35 and 39}
\newcommand{\radiologynetHomogeneousClusters}{2, 3, 8, 25, 43 and 44}

\begin{figure}[!tb]
    \centering
    \includegraphics[width=\linewidth]{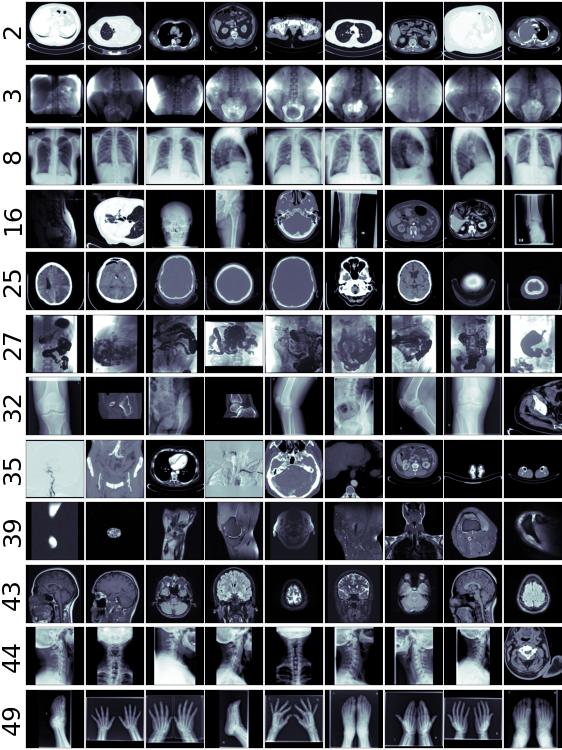}
    \caption{Randomly selected images from select clusters. Cluster indices are indicated to the left of each row.}
    \label{fig:cluster_snippets_mosaic}
\end{figure}

\begin{figure*}[!tb]
    \centering
    \includegraphics[width=\linewidth]{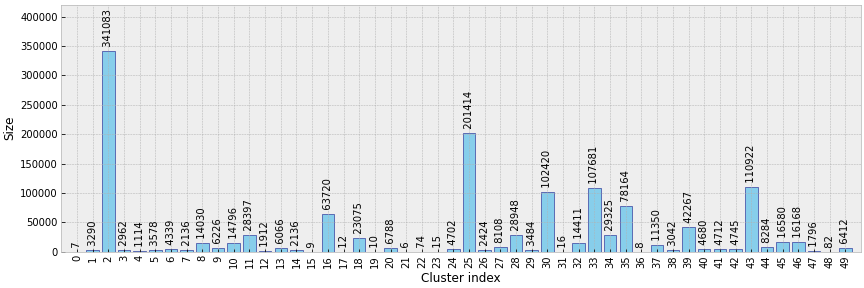}
    \caption{Sizes of obtained groups in the labelled RadiologyNET dataset.}
    \label{fig:sizes_of_obtained_groups}
\end{figure*}

\begin{figure}[!tb]
    \centering
    \includegraphics[width=\linewidth]{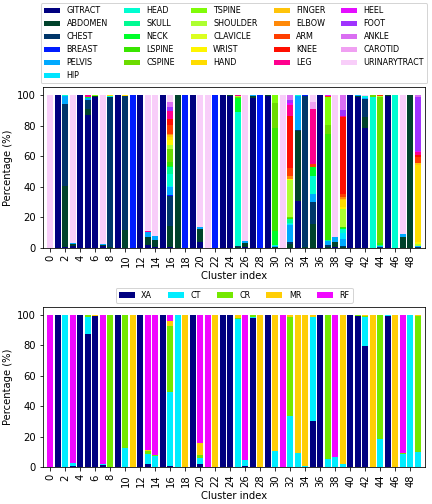}
    \caption{Grouping quality regarding modality and examined body part, for the labelled RadiologyNET dataset. The first image shows how homogeneous the clusters are when observing the body part, while the second one shows the different modalities in each cluster.}
    \label{fig:dataset_grouping_quality_by_mod_and_bpe_entire_radiologynet}
\end{figure}

From the original dataset described in~\ref{sec:datpp} we extracted a set of {\radiologynetSize} images which fit the aforementioned criteria related to image, diagnosis and DICOM tag policies. The chosen labelling algorithm was used to cluster this more extensive set of DICOM instances into 50 groups. As is shown in Fig.~\ref{fig:sizes_of_obtained_groups}, the obtained groups varied in size, with the largest one having {\radiologynetLargestClusterSize} instances (data points), and the smallest one comprising of only {\radiologynetSmallestClusterSize} instances.
Such small clusters can be considered to contain anomalies which do not fit into other groups. 
Fig.~\ref{fig:dataset_grouping_quality_by_mod_and_bpe_entire_radiologynet} shows the quality of grouping throughout all of the clusters. One can see how the quality of grouping corresponds to the one shown in Fig.~\ref{fig:dataset_grouping_quality_by_mod_and_bpe_pvdbow_ae_cae}b), indicating that using new data, which was previously unseen by the labelling algorithm, did not significantly influence the quality of clusters. Groups which were heterogeneous on the smaller set ({\radiologynetHeterogeneousClusters}) remained the same after labelling the larger set, and the same applies to homogeneous clusters such as {\radiologynetHomogeneousClusters}.
Random samples of images from these (and other) clusters can be seen in Fig.~\ref{fig:cluster_snippets_mosaic}. 

Next, we consider the quality of obtained groups (Fig.~\ref{fig:dataset_grouping_quality_by_mod_and_bpe_entire_radiologynet}). It can be seen that almost all clusters show a high level of homogeneity when considering their imaging modality. On the other hand, body part homogeneity shows how nearby anatomic regions are often grouped together. This is especially prominent in the torso region, where it is difficult to accurately discern the exact border between the abdomen, the gastrointestinal tract and the pelvis (cluster 33), as well as the abdomen and the chest (cluster 2). Also, it is not unusual that a single study contains multiple parts of the torso, for example, an MR image capturing both the abdomen and the pelvis. The same applies to groupings of the spine, where different spinal parts are often assigned to the same group (clusters 30, 37). Images capturing the extremities (hands and feet) were also grouped together in cluster 49, despite being visually dissimilar from one another and showing different anatomical regions. On the other hand, clusters {\radiologynetHeterogeneousClusters} showed evidence of containing anomalies; they contained images depicting non-connected anatomical regions (leg, abdomen, head, urinary tract...).

\section{Conclusion}\label{sec:con}
In this paper, we deal with the problem of building a precursor to a sizeable annotated dataset of medical radiology images, using unsupervised machine learning methods to discover useful patterns by combining three data sources: DICOM metadata, images and narrative diagnoses. The purpose of the dataset is to create an ImageNet counterpart for training standard deep learning classification model architectures tuned to medical radiology imaging tasks. Obtained clustering rules exhibit good homogeneity regarding the imaging modality and the anatomic region on a representative data subset.

The final evaluation of the RadiologyNET annotation system will involve assessing TL models on open challenges in medical radiology image processing. This testing will provide a comprehensive and unbiased evaluation of the system's performance and enable a direct comparison with other state-of-the-art TL techniques; and will also assess the system's robustness, accuracy, and how well it can generalise problems. One thing that must be pointed out is the possibility of reduced TL performance due to the distribution shift and the currently limited number of distinct classes. While we aim to provide comprehensive fine-grained coverage of various imaging modalities and protocols, it is important to note that not all modalities and protocols are currently included in the annotated dataset, nor is the number of distinct visual categories large enough to make it comparable to ImageNet. Hence, these limitations may potentially impact TL efficiency in practical applications.

The described approach is fully unsupervised, whereas using supervised methods remains a topic of future work. Leveraging modality and examined body part to train supervised feature extractors could possibly benefit the overall quality of the labelled dataset. 
On another note, textual diagnoses proved to be efficient in grouping medical images, achieving a high $S$ score and having the resulting cluster be homogeneous regarding the anatomical region they are depicting. This work could be expanded upon by exploring sophisticated natural language processing (NLP) methods such as GPT or BERT and using sentence encoders to get more accurate diagnoses embeddings.

In summary, in addition to documenting the birth of the RadiologyNET dataset, this study provides insights into automating the labelling process of DICOM data, highlighting the challenges and achievements in grouping them based on anatomical region and imaging modality. The findings contribute to a better understanding of the limitations and potential improvements in automated labelling algorithms for medical imaging datasets. Future research can build upon these findings to refine and enhance the grouping process, ultimately aiding in more accurate and meaningful analysis of medical images.

\section*{Summary table}
\subsection*{What was already known on the topic}
\begin{itemize}
    \item Medical data annotation is time consuming, expensive, demanding and subject to the level of annotators' knowledge and bias.
    \item Unsupervised machine learning techniques can be used to extract useful features from various data sources, by exploiting useful patterns in the data. 
\end{itemize}
\subsection*{What this study added to our knowledge}
\begin{itemize}
    \item An extensive study of applicable unsupervised feature extractors was completed for three different typical data types present in PACS databases -- images, DICOM tags and narrative diagnoses -- and best practice for each data source was established.
    \item An in-depth methodological approach for large-scale medical data grouping has been explored and the complete procedure has been described in great detail.
    \item Unsupervised machine learning techniques can be used for annotating semantically similar medical radiology images from multimodal data sources.
\end{itemize}

\section*{CRediT authorship contribution statement}
\textbf{Mateja Napravnik}: Conceptualisation, Methodology, Software, Writing – original draft, Writing – review \& editing, Visualisation.
\textbf{Franko Hržić}: Conceptualisation, Methodology, Software, Writing – original draft, Writing – review \& editing, Visualisation.
\textbf{Sebastian Tschauner}: Methodology, Writing – review \& editing, Supervision. 
\textbf{Ivan Štajduhar}: Conceptualisation, Resources, Writing – review \& editing, Supervision, Project administration, Funding acquisition.

\section*{Declaration of competing interest}
The authors declare that they have no known competing financial interests or personal relationships that could have appeared to influence the work reported in this paper.

\section*{Data availability}
The dataset used in this study is not available for sharing, according to the current Ethics Committee approval. However, the entire code used for the experiments is available at \url{https://github.com/AIlab-RITEH/RadiologyNET-public}.
 
\section*{Funding}
This work has been fully supported by the Croatian Science Foundation [grant number IP-2020-02-3770].

\bibliographystyle{elsarticle-num} 
\bibliography{bibliography}

\begin{thebibliography}{10}
\expandafter\ifx\csname url\endcsname\relax
  \def\url#1{\texttt{#1}}\fi
\expandafter\ifx\csname urlprefix\endcsname\relax\def\urlprefix{URL }\fi
\expandafter\ifx\csname href\endcsname\relax
  \def\href#1#2{#2} \def\path#1{#1}\fi

\bibitem{Nagy:2022:grazped_dataset}
E.~Nagy, M.~Janisch, F.~Hr{\v{z}}i{\'{c}}, E.~Sorantin, S.~Tschauner, {A
  pediatric wrist trauma X-ray dataset (GRAZPEDWRI-DX) for machine learning},
  Scientific Data 9~(1) (2022) 222.
\newblock \href {https://doi.org/10.1038/s41597-022-01328-z}
  {\path{doi:10.1038/s41597-022-01328-z}}.

\bibitem{Irvin:2019:CheXpert_dataset}
J.~Irvin, P.~Rajpurkar, M.~Ko, Y.~Yu, S.~Ciurea{-}Ilcus, C.~Chute, H.~Marklund,
  B.~Haghgoo, R.~L. Ball, K.~S. Shpanskaya, J.~Seekins, D.~A. Mong, S.~S.
  Halabi, J.~K. Sandberg, R.~Jones, D.~B. Larson, C.~P. Langlotz, B.~N. Patel,
  M.~P. Lungren, A.~Y. Ng, {CheXpert: {A} Large Chest Radiograph Dataset with
  Uncertainty Labels and Expert Comparison}, CoRR abs/1901.07031 (2019).
\newblock \href {http://arxiv.org/abs/1901.07031} {\path{arXiv:1901.07031}}.

\bibitem{rajpurkar:2018:mura_dataset}
P.~Rajpurkar, J.~Irvin, A.~Bagul, D.~Ding, T.~Duan, H.~Mehta, B.~Yang, K.~Zhu,
  D.~Laird, R.~L. Ball, C.~Langlotz, K.~Shpanskaya, M.~P. Lungren, A.~Y. Ng,
  {MURA: Large Dataset for Abnormality Detection in Musculoskeletal
  Radiographs} (2018).
\newblock \href {http://arxiv.org/abs/1712.06957} {\path{arXiv:1712.06957}}.

\bibitem{weiss2016survey}
K.~Weiss, T.~M. Khoshgoftaar, D.~Wang, {A survey of transfer learning}, Journal
  of Big Data 3~(1) (may 2016).
\newblock \href {https://doi.org/10.1186/s40537-016-0043-6}
  {\path{doi:10.1186/s40537-016-0043-6}}.

\bibitem{pan2010survey}
S.~J. Pan, Q.~Yang, {A survey on transfer learning}, IEEE Transactions on
  knowledge and data engineering 22~(10) (2010) 1345--1359.

\bibitem{krizhevsky2017imagenet}
A.~Krizhevsky, I.~Sutskever, G.~E. Hinton, {{ImageNet} classification with deep
  convolutional neural networks}, Communications of the {ACM} 60~(6) (2017)
  84--90.
\newblock \href {https://doi.org/10.1145/3065386} {\path{doi:10.1145/3065386}}.

\bibitem{paszke2019pytorch}
A.~Paszke, S.~Gross, F.~Massa, A.~Lerer, J.~Bradbury, G.~Chanan, T.~Killeen,
  Z.~Lin, N.~Gimelshein, L.~Antiga, et~al., Pytorch: An imperative style,
  high-performance deep learning library, Advances in neural information
  processing systems 32 (2019).

\bibitem{abadi2016tensorflow}
M.~Abadi, A.~Agarwal, P.~Barham, E.~Brevdo, Z.~Chen, C.~Citro, G.~S. Corrado,
  A.~Davis, J.~Dean, M.~Devin, S.~Ghemawat, I.~Goodfellow, A.~Harp, G.~Irving,
  M.~Isard, Y.~Jia, R.~Jozefowicz, L.~Kaiser, M.~Kudlur, J.~Levenberg, D.~Mane,
  R.~Monga, S.~Moore, D.~Murray, C.~Olah, M.~Schuster, J.~Shlens, B.~Steiner,
  I.~Sutskever, K.~Talwar, P.~Tucker, V.~Vanhoucke, V.~Vasudevan, F.~Viegas,
  O.~Vinyals, P.~Warden, M.~Wattenberg, M.~Wicke, Y.~Yu, X.~Zheng, {TensorFlow:
  Large-Scale Machine Learning on Heterogeneous Distributed Systems} (2016).
\newblock \href {https://doi.org/10.48550/ARXIV.1603.04467}
  {\path{doi:10.48550/ARXIV.1603.04467}}.

\bibitem{raghu2019transfusion}
M.~Raghu, C.~Zhang, J.~Kleinberg, S.~Bengio,
  \href{https://proceedings.neurips.cc/paper_files/paper/2019/file/eb1e78328c46506b46a4ac4a1e378b91-Paper.pdf}{{Transfusion:
  Understanding Transfer Learning for Medical Imaging}} (2019).
\newline\urlprefix\url{https://proceedings.neurips.cc/paper_files/paper/2019/file/eb1e78328c46506b46a4ac4a1e378b91-Paper.pdf}

\bibitem{Alzubaidi:2021:BigData}
L.~Alzubaidi, J.~Zhang, A.~J. Humaidi, A.~Al-Dujaili, Y.~Duan, O.~Al-Shamma,
  J.~Santamar{\'i}a, M.~A. Fadhel, M.~Al-Amidie, L.~Farhan,
  \href{https://doi.org/10.1186/s40537-021-00444-8}{{Review of deep learning:
  concepts, CNN architectures, challenges, applications, future directions}},
  Journal of Big Data 8~(1) (2021) 53.
\newblock \href {https://doi.org/10.1186/s40537-021-00444-8}
  {\path{doi:10.1186/s40537-021-00444-8}}.
\newline\urlprefix\url{https://doi.org/10.1186/s40537-021-00444-8}

\bibitem{alzubaidi2020towards}
L.~Alzubaidi, M.~A. Fadhel, O.~Al-Shamma, J.~Zhang, J.~Santamar{\'{\i}}a,
  Y.~Duan, S.~R. Oleiwi, {Towards a Better Understanding of Transfer Learning
  for Medical Imaging: A Case Study}, Applied Sciences 10~(13) (2020) 4523.
\newblock \href {https://doi.org/10.3390/app10134523}
  {\path{doi:10.3390/app10134523}}.

\bibitem{mustafa2021supervised}
B.~Mustafa, A.~Loh, J.~Freyberg, P.~MacWilliams, M.~Wilson, S.~M. McKinney,
  M.~Sieniek, J.~Winkens, Y.~Liu, P.~Bui, S.~Prabhakara, U.~Telang,
  A.~Karthikesalingam, N.~Houlsby, V.~Natarajan, {Supervised Transfer Learning
  at Scale for Medical Imaging} (2021).
\newblock \href {https://doi.org/10.48550/ARXIV.2101.05913}
  {\path{doi:10.48550/ARXIV.2101.05913}}.

\bibitem{enders2022applied-missing-data}
C.~K. Enders, {Applied missing data analysis}, Guilford Publications, 2022.

\bibitem{bhaskaran2014differenceMCAR}
K.~Bhaskaran, L.~Smeeth, {What is the difference between missing completely at
  random and missing at random?}, International Journal of Epidemiology 43~(4)
  (2014) 1336--1339.
\newblock \href {https://doi.org/10.1093/ije/dyu080}
  {\path{doi:10.1093/ije/dyu080}}.

\bibitem{Emmanuel2021missing_data_ml}
T.~Emmanuel, T.~Maupong, D.~Mpoeleng, T.~Semong, B.~Mphago, O.~Tabona, {A
  survey on missing data in machine learning}, Journal of Big Data 8~(1) (oct
  2021).
\newblock \href {https://doi.org/10.1186/s40537-021-00516-9}
  {\path{doi:10.1186/s40537-021-00516-9}}.

\bibitem{stekhoven2012missforest}
D.~J. Stekhoven, P.~Bühlmann, {MissForest}{\textemdash}non-parametric missing
  value imputation for mixed-type data, Bioinformatics 28~(1) (2011) 112--118.
\newblock \href {https://doi.org/10.1093/bioinformatics/btr597}
  {\path{doi:10.1093/bioinformatics/btr597}}.

\bibitem{Tang2017RF_Missingdata}
F.~Tang, H.~Ishwaran, {Random forest missing data algorithms}, Statistical
  Analysis and Data Mining: The {ASA} Data Science Journal 10~(6) (2017)
  363--377.
\newblock \href {https://doi.org/10.1002/sam.11348}
  {\path{doi:10.1002/sam.11348}}.

\bibitem{larobina2014medical}
M.~Larobina, L.~Murino, {Medical Image File Formats}, Journal of Digital
  Imaging 27~(2) (2013) 200--206.
\newblock \href {https://doi.org/10.1007/s10278-013-9657-9}
  {\path{doi:10.1007/s10278-013-9657-9}}.

\bibitem{dicom_standard}
{DICOM Standards Committee}, { DICOM Standard },
  https://www.dicomstandard.org/, accessed on 5 Apr 2023 (2023).

\bibitem{Yang2019sentence_encoders}
Y.~Yang, D.~Cer, A.~Ahmad, M.~Guo, J.~Law, N.~Constant, G.~H. Abrego, S.~Yuan,
  C.~Tar, Y.-H. Sung, B.~Strope, R.~Kurzweil, {Multilingual Universal Sentence
  Encoder for Semantic Retrieval} (2019).
\newblock \href {https://doi.org/10.48550/ARXIV.1907.04307}
  {\path{doi:10.48550/ARXIV.1907.04307}}.

\bibitem{openai:2023:gpt4}
OpenAI, {GPT-4 Technical Report} (2023).
\newblock \href {http://arxiv.org/abs/2303.08774} {\path{arXiv:2303.08774}}.

\bibitem{ljubevsic:2007:stemmer}
N.~Ljube{\v{s}}i{\'c}, D.~Boras, O.~Kubelka, {Retrieving information in
  Croatian: Building a simple and efficient rule-based stemmer}, in: The Future
  of Information Sciences" (INFuture 2007) : Digital information and heritage,
  Odsjek za informacijske znanosti, Filozofski fakultet, Zagreb, 2007, pp.
  313--320.

\bibitem{anderson1999lapack}
E.~Anderson, Z.~Bai, C.~Bischof, L.~S. Blackford, J.~Demmel, J.~Dongarra,
  J.~Du~Croz, A.~Greenbaum, S.~Hammarling, A.~McKenney, et~al., {LAPACK users'
  guide}, SIAM, 1999.

\bibitem{lehoucq1998arpack}
R.~B. Lehoucq, D.~C. Sorensen, C.~Yang, {ARPACK users' guide: solution of
  large-scale eigenvalue problems with implicitly restarted Arnoldi methods},
  SIAM, 1998.

\bibitem{martinsson2011pcarandomized}
P.-G. Martinsson, V.~Rokhlin, M.~Tygert, {A randomized algorithm for the
  decomposition of matrices}, Applied and Computational Harmonic Analysis
  30~(1) (2011) 47--68.
\newblock \href {https://doi.org/10.1016/j.acha.2010.02.003}
  {\path{doi:10.1016/j.acha.2010.02.003}}.

\bibitem{abdi2010pca}
H.~Abdi, L.~J. Williams, {Principal component analysis}, Wiley
  Interdisciplinary Reviews: Computational Statistics 2~(4) (2010) 433--459.
\newblock \href {https://doi.org/10.1002/wics.101}
  {\path{doi:10.1002/wics.101}}.

\bibitem{napravnik2022autoencoders}
M.~Napravnik, R.~Baždarić, D.~Miletić, F.~Hržić, S.~Tschauner, M.~Mamula,
  I.~Štajduhar, {Using Autoencoders to Reduce Dimensionality of DICOM
  Metadata}, in: 2022 International Conference on Electrical, Computer,
  Communications and Mechatronics Engineering (ICECCME), 2022, pp. 1--6.
\newblock \href {https://doi.org/10.1109/ICECCME55909.2022.9988310}
  {\path{doi:10.1109/ICECCME55909.2022.9988310}}.

\bibitem{agarap2018relu}
A.~F. Agarap, {Deep Learning using Rectified Linear Units (ReLU)} (2018).
\newblock \href {https://doi.org/10.48550/ARXIV.1803.08375}
  {\path{doi:10.48550/ARXIV.1803.08375}}.

\bibitem{tajbakhsh2020med_image_seg_methods}
N.~Tajbakhsh, L.~Jeyaseelan, Q.~Li, J.~N. Chiang, Z.~Wu, X.~Ding, {Embracing
  imperfect datasets: A review of deep learning solutions for medical image
  segmentation}, Medical Image Analysis 63 (2020) 101693.
\newblock \href {https://doi.org/10.1016/j.media.2020.101693}
  {\path{doi:10.1016/j.media.2020.101693}}.

\bibitem{ronneberger2015unet}
O.~Ronneberger, P.~Fischer, T.~Brox, {U-Net: Convolutional Networks for
  Biomedical Image Segmentation}, in: Lecture Notes in Computer Science,
  Springer International Publishing, 2015, pp. 234--241.
\newblock \href {https://doi.org/10.1007/978-3-319-24574-4_28}
  {\path{doi:10.1007/978-3-319-24574-4_28}}.

\bibitem{alom2018r2unet}
M.~Z. Alom, M.~Hasan, C.~Yakopcic, T.~M. Taha, V.~K. Asari, {Recurrent Residual
  Convolutional Neural Network based on U-Net (R2U-Net) for Medical Image
  Segmentation} (2018).
\newblock \href {https://doi.org/10.48550/ARXIV.1802.06955}
  {\path{doi:10.48550/ARXIV.1802.06955}}.

\bibitem{oktay2018attentionunet}
O.~Oktay, J.~Schlemper, L.~L. Folgoc, M.~Lee, M.~Heinrich, K.~Misawa, K.~Mori,
  S.~McDonagh, N.~Y. Hammerla, B.~Kainz, B.~Glocker, D.~Rueckert, {Attention
  U-Net: Learning Where to Look for the Pancreas} (2018).
\newblock \href {https://doi.org/10.48550/ARXIV.1804.03999}
  {\path{doi:10.48550/ARXIV.1804.03999}}.

\bibitem{Mikolov:2013:cbow}
T.~Mikolov, K.~Chen, G.~Corrado, J.~Dean, {Efficient Estimation of Word
  Representations in Vector Space}, publisher = {arXiv} (2013).
\newblock \href {https://doi.org/10.48550/ARXIV.1301.3781}
  {\path{doi:10.48550/ARXIV.1301.3781}}.

\bibitem{Dang:2020:TDIDF}
N.~C. Dang, M.~N. Moreno-García, F.~De~la Prieta, {Sentiment Analysis Based on
  Deep Learning: A Comparative Study}, Electronics 9~(3) (2020).
\newblock \href {https://doi.org/10.3390/electronics9030483}
  {\path{doi:10.3390/electronics9030483}}.

\bibitem{Mikolov:2014:Doc2Vec}
Q.~Le, T.~Mikolov,
  \href{https://proceedings.mlr.press/v32/le14.html}{{Distributed
  Representations of Sentences and Documents}}, in: E.~P. Xing, T.~Jebara
  (Eds.), Proceedings of the 31st International Conference on Machine Learning,
  Vol. 32(2) of Proceedings of Machine Learning Research, PMLR, Bejing, China,
  2014, pp. 1188--1196.
\newline\urlprefix\url{https://proceedings.mlr.press/v32/le14.html}

\bibitem{KHATTAK:2019:survey_text}
F.~K. Khattak, S.~Jeblee, C.~Pou-Prom, M.~Abdalla, C.~Meaney, F.~Rudzicz, {A
  survey of word embeddings for clinical text}, Journal of Biomedical
  Informatics 100 (2019) 100057.
\newblock \href {https://doi.org/10.1016/j.yjbinx.2019.100057}
  {\path{doi:10.1016/j.yjbinx.2019.100057}}.

\bibitem{Kowsar:2019:survey_text2}
K.~Kowsari, K.~Jafari~Meimandi, M.~Heidarysafa, S.~Mendu, L.~Barnes, D.~Brown,
  {Text Classification Algorithms: A Survey}, Information 10~(4) (2019).
\newblock \href {https://doi.org/10.3390/info10040150}
  {\path{doi:10.3390/info10040150}}.

\bibitem{Li:2022:survey_text_3}
Q.~Li, H.~Peng, J.~Li, C.~Xia, R.~Yang, L.~Sun, P.~S. Yu, L.~He, {A Survey on
  Text Classification: From Traditional to Deep Learning}, ACM Trans. Intell.
  Syst. Technol. 13~(2) (apr 2022).
\newblock \href {https://doi.org/10.1145/3495162} {\path{doi:10.1145/3495162}}.

\bibitem{DBLP:2018:BERT}
J.~Devlin, M.~Chang, K.~Lee, K.~Toutanova, {{BERT:} Pre-training of Deep
  Bidirectional Transformers for Language Understanding}, CoRR abs/1810.04805
  (2018).
\newblock \href {http://arxiv.org/abs/1810.04805} {\path{arXiv:1810.04805}},
  \href {https://doi.org/10.48550/ARXIV.1810.04805}
  {\path{doi:10.48550/ARXIV.1810.04805}}.

\bibitem{DBLP:2020:GPT}
T.~B. Brown, B.~Mann, N.~Ryder, M.~Subbiah, J.~Kaplan, P.~Dhariwal,
  A.~Neelakantan, P.~Shyam, G.~Sastry, A.~Askell, S.~Agarwal,
  A.~Herbert{-}Voss, G.~Krueger, T.~Henighan, R.~Child, A.~Ramesh, D.~M.
  Ziegler, J.~Wu, C.~Winter, C.~Hesse, M.~Chen, E.~Sigler, M.~Litwin, S.~Gray,
  B.~Chess, J.~Clark, C.~Berner, S.~McCandlish, A.~Radford, I.~Sutskever,
  D.~Amodei, {Language Models are Few-Shot Learners}, CoRR abs/2005.14165
  (2020).
\newblock \href {http://arxiv.org/abs/2005.14165} {\path{arXiv:2005.14165}}.

\bibitem{elkan2003kmeans}
C.~Elkan, {Using the triangle inequality to accelerate k-means}, in:
  Proceedings of the 20th international conference on Machine Learning
  (ICML-03), 2003, pp. 147--153.

\bibitem{rdusseeun1987clusteringMedoids}
L.~Rdusseeun, P.~Kaufman, {Clustering by means of medoids}, in: Proceedings of
  the statistical data analysis based on the L1 norm conference, neuchatel,
  switzerland, Vol.~31, 1987, pp. 405--416.

\bibitem{Kvaalseth2017nmi}
T.~Kv{\aa}lseth, {On Normalized Mutual Information: Measure Derivations and
  Properties}, Entropy 19~(11) (2017) 631.
\newblock \href {https://doi.org/10.3390/e19110631}
  {\path{doi:10.3390/e19110631}}.

\bibitem{thorndike1953elbow_belongs}
R.~L. Thorndike, {Who belongs in the family?}, Psychometrika 18~(4) (1953)
  267--276.
\newblock \href {https://doi.org/10.1007/bf02289263}
  {\path{doi:10.1007/bf02289263}}.

\bibitem{satopaa2011finding_knee}
V.~Satopaa, J.~Albrecht, D.~Irwin, B.~Raghavan, {Finding a "Kneedle" in a
  Haystack: Detecting Knee Points in System Behavior}, in: 2011 31st
  International Conference on Distributed Computing Systems Workshops, {IEEE},
  2011, pp. 166--171.
\newblock \href {https://doi.org/10.1109/icdcsw.2011.20}
  {\path{doi:10.1109/icdcsw.2011.20}}.

\bibitem{rosenberg2007v}
A.~Rosenberg, J.~Hirschberg, {V-measure: A conditional entropy-based external
  cluster evaluation measure}, in: Proceedings of the 2007 joint conference on
  empirical methods in natural language processing and computational natural
  language learning (EMNLP-CoNLL), 2007, pp. 410--420.

\bibitem{radford2021clip}
A.~Radford, J.~W. Kim, C.~Hallacy, A.~Ramesh, G.~Goh, S.~Agarwal, G.~Sastry,
  A.~Askell, P.~Mishkin, J.~Clark, et~al., {Learning transferable visual models
  from natural language supervision}, in: International conference on machine
  learning, PMLR, 2021, pp. 8748--8763.

\bibitem{vreeman2018loinc}
D.~J. Vreeman, S.~Abhyankar, K.~C. Wang, C.~Carr, B.~Collins, D.~L. Rubin,
  C.~P. Langlotz, {The {LOINC} {RSNA} radiology playbook - a unified
  terminology for radiology procedures}, Journal of the American Medical
  Informatics Association 25~(7) (2018) 885--893.
\newblock \href {https://doi.org/10.1093/jamia/ocy053}
  {\path{doi:10.1093/jamia/ocy053}}.

\bibitem{Stajduhar2021-biomesip}
I.~{\v{S}}tajduhar, T.~Manojlovi{\'c}, F.~Hr{\v{z}}i{\'c}, M.~Napravnik,
  G.~Glava{\v{s}}, M.~Milani{\v{c}}, S.~Tschauner,
  M.~Mamula~Sara{\v{c}}evi{\'c}, D.~Mileti{\'c}, {Analysing Large Repositories
  of Medical Images}, in: Bioengineering and Biomedical Signal and Image
  Processing, Springer International Publishing, 2021, pp. 179--193.
\newblock \href {https://doi.org/10.1007/978-3-030-88163-4_17}
  {\path{doi:10.1007/978-3-030-88163-4_17}}.

\bibitem{mildenberger2002introduction}
P.~Mildenberger, M.~Eichelberg, E.~Martin, {Introduction to the {DICOM}
  standard}, European Radiology 12~(4) (2001) 920--927.
\newblock \href {https://doi.org/10.1007/s003300101100}
  {\path{doi:10.1007/s003300101100}}.

\bibitem{thompson:2002:LUT}
S.~K. Thompson, C.~E. Willis, K.~T. Krugh, S.~J. Shepard, K.~W. McEnery,
  {Implementing the {DICOM} Grayscale Standard Display Function for Mixed Hard-
  and Soft-Copy Operations}, Journal of Digital Imaging 15~(0) (2002) 27--32.
\newblock \href {https://doi.org/10.1007/s10278-002-5026-9}
  {\path{doi:10.1007/s10278-002-5026-9}}.

\bibitem{Hrzic:2022:export}
F.~Hržić, M.~Napravnik, R.~Baždarić, I.~Štajduhar, M.~Mamula, D.~Miletić,
  S.~Tschauner, {Estimation of Missing Parameters for DICOM to 8-bit X-ray
  Image Export}, in: 2022 International Conference on Electrical, Computer,
  Communications and Mechatronics Engineering (ICECCME), 2022, pp. 1--6.
\newblock \href {https://doi.org/10.1109/ICECCME55909.2022.9988674}
  {\path{doi:10.1109/ICECCME55909.2022.9988674}}.

\end{thebibliography}

\appendix

\section{Variable description}
Variables used in the text are listed and briefly described in Table~\ref{tab:vardesc}.

\begin{table}[!b]
    \caption{A list and a brief description of each variable used in the text.}
    \centering
    \footnotesize
    \begin{tabular}{c|p{5cm}}
        \textbf{Variable name} & \textbf{Variable description} \\
        \hline
        $\kappa$ & Number of clusters. \\
        $\hat{y}$ & Predicted cluster label \\
        $\hat{y}^{(i)}$ & Predicted cluster label of instance $i$\\
        $y^{(i)}_t$ & Observed target variable $t$ value of instance $i$ ($y_{B}$ or $y_{M}$). \\
        $\hat{y}^{(i)}_t$ & Predicted target variable $t$ value of instance $i$ ($\hat{y}_{B}$ or $\hat{y}_{M}$). \\
        $x^{(i)}$ & Input data instance $i$ (all sources). \\
        $x^{(i)}_{I}$ & Input data instance $i$ image. \\
        $x^{(i)}_{T}$ & Input data instance $i$ DICOM tags. \\
        $x^{(i)}_{D}$ & Input data instance $i$ diagnosis. \\
        $f\big(x^{(i)}\big)$ & Input data instance $i$ embedding (all sources). \\
        $f\big(x^{(i)}_{I}\big)$ & Input data instance $i$ image embedding. \\
        $f\big(x^{(i)}_{T}\big)$ & Input data instance $i$ DICOM tags embedding. \\
        $f\big(x^{(i)}_{D}\big)$ & Input data instance $i$ diagnosis embedding. \\

        $NMI_M$ & NMI regarding Modality. \\
        $NMI_B$ & NMI regarding BodyPartExamined. \\
        $HS_M$ & HS regarding Modality. \\
        $HS_B$ & HS regarding BodyPartExamined. \\
        $S$ & The harmonic mean of $HS_B$, $HS_M$, $NMI_M$ and $NMI_B$ which will be used to select diagnosis/image models. \\

        $D_I$ & The mean distance of image embeddings across all clusters. \\
        $D_D$ & The mean distance of diagnoses embeddings across all clusters. \\ 
        $D_{score}$ & The harmonic mean of $D_I$ and $D_D.$ \\
    \end{tabular}
    \label{tab:vardesc}
\end{table}

\section{Extraction of experimental data subset}\label{sec:datpp}
The material presented here contains the details that complement the text in section~\ref{ssec:data_cleanup}.
The dataset consists of approximately $2$ million unique exams completed between 2008 and 2017 and performed through standard clinical practice at Clinical Hospital Centre (CHC) Rijeka. It was gathered retrospectively in 2017. The data was anonymised during the extraction process to remove all sensitive information. We obtained approval from the competent Ethics Committee to collect and process the data for this purpose. The obtained approval mandates that the data remains private in its current form.

\begin{figure}
    \centering
    \includegraphics[width=0.6\linewidth]{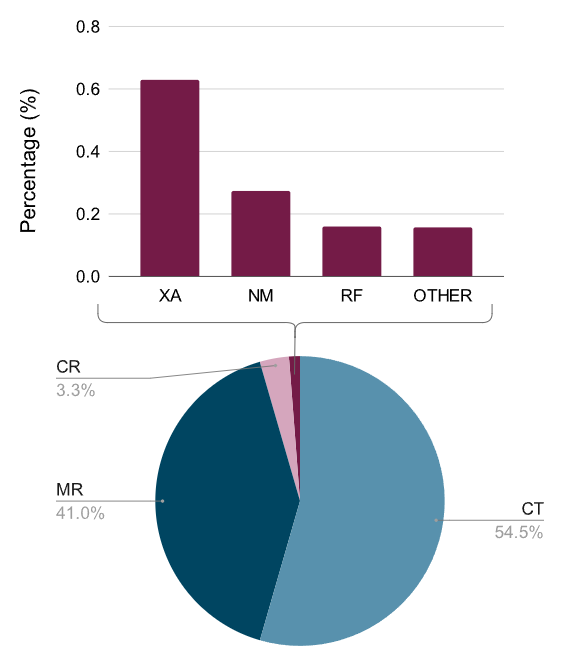}
    \caption{Distribution of modalities across the entire $25$ million DICOM files.}
    \label{fig:dist_modalities_all_available_data}
\end{figure}

Each exam could result in a respective diagnosis and at least one (often more than one) DICOM file, which was then stored at CHC Rijeka's Picture Archiving and Communication System (PACS). The total number of DICOM files obtained from CHC Rijeka PACS reached approximately $25$ million~\cite{Stajduhar2021-biomesip}. The most prevalent modalities in the dataset were CT, MR, XA, NM and RF, as is shown in Fig.~\ref{fig:dist_modalities_all_available_data}.

Not all exams resulted in a recorded diagnosis, meaning that diagnoses of some of the performed exams were empty or null. This was the case in roughly $6.96\%$ of data, while $93.03\%$ contained a non-null, non-empty value. Upon further inspection, we discovered that some of the non-empty diagnoses were less than {\lowestNrOfCharactersInDiagnoses} characters long. These were presumed to be anomalies as such short diagnoses could seldom carry useful information. Empty diagnoses and diagnoses which had less than {\lowestNrOfCharactersInDiagnoses} characters were excluded from the used dataset.

From the original dataset we sampled a subset of {\numberOfImagesInSubset} DICOM files and adjoined textual diagnoses.
During the sampling process, special attention was given to include as many complete diagnoses as possible.
Namely, an analysis of the available data revealed that there were examinations which resulted in more than 1000 different DICOM files due to multiple projections or views used by radiologists. To avoid a disproportion between the number of distinct diagnoses and DICOM files in the subset, a threshold of {\maxNrOfImagesInExam} files was chosen. This means that all exams which resulted in more than {\maxNrOfImagesInExam} DICOM files were eliminated outright.
The sampled subset contained images acquired in $5$ different modalities: CT, MR, CR, XA and RF. Although initially included in the subset, images recorded in the NM modality were removed because they were often associated with examinations whose diagnoses were empty, short, or otherwise uninformative.

\section{DICOM Tags extraction and imputation}~\label{AP:DICOM TAGS}
The material presented here, where we address several groups of problems related to DICOM tags, contains the details that complement the text in section~\ref{sec:dcmtag}.
The first encountered problem was \textbf{DICOM tags with missing BPE}, which contained an empty value in {\bpeEmptyPercent} cases. On the other hand, tags such as \textit{ProtocolName}, \textit{StudyDescription} and \textit{RequestedProcedureDescription} faired better, having empty values in only {\ProtocolNameEmptyPercentage}, {\StudyDescEmptyPercentage} and {\ReqProcedureDescEmptyPercentage} of instances, respectively.
Wherever \textit{BodyPartExamined} was empty, at least one of the mentioned tags contained a value from which one can infer the examined body part, which is why these three particular tags were chosen.
In order to solve the missing values for \textit{BodyPartExamined} tag, there were {\nrOfRegexWritten} regular expressions written, which contained rules for imputing \textit{BodyPartExamined} from the \textit{ProtocolName}, \textit{StudyDescription} and \textit{RequestedProcedureDescription} tags. These were also written in a way that account for possible typographical errors (e.g.~\textit{torax} and \textit{thorax}), multiple languages used by physicians (e.g.~Latin: \textit{calcaneus}; English: \textit{heel bone}; and Croatian: \textit{petna kost}), possible abbreviations (e.g.~\textit{c-spine}, \textit{c\_spine}, \textit{cspine} and \textit{cervical spine}), and which procedures impact which body part (e.g.~\textit{chemoembolization} is tied to the liver, which is a part of the urinary tract).  These rules were written under a radiologist's guidance, as there is no straightforward ruleset for perfect BPE mapping. 

The final result of BPE imputation (based on knowledge, decision rules, and regular expressions) was manifested in a jump from {\bpeFullPercent} to $100\%$ non-empty instances. However, we should note that there was still a possibility of erroneously imputing \textit{BodyPartExamined} from other tags. Specifically, DICOM tags \textit{StudyDescription}, \textit{ProtocolName} and \textit{RequestedProcedureDescription} are input manually by a performing physician. As such, other than typographical errors, it is possible that other types of errors could lead to mislabelling of a body part that was not accounted for. However, these cases were presumed to be anomalies and only present in a few DICOM files.

The next group of tags needing additional care are stringified arrays - \textbf{DICOM tags with multiple values}. Namely, DICOM tags can contain multiple values, for example, \textit{ImageType} and \textit{WindowCenter}. Such tags were parsed from a single stringified array-like tag into multiple tags, which resulted in \textit{ImageType} dissolving into \textit{ImageType0} and \textit{ImageType1}, etc.

Another group of DICOM tag problems was \textbf{selection of appropriate DICOM tags}. 
There were {\initialNrOfDcmTags} different DICOM tags that appeared at least once in the whole subset. However, many proved to be uninformative due to either being empty in most instances or having only one distinct value. A fill rate threshold was imposed on each tag, and each tag with less than {\dcmTagFillrateThreshold} non-empty values was removed. Furthermore, all DICOM tags with less than {\dcmTagUniqueValuesThreshold} distinct values were discarded, along with tags containing unique identifiers, such as \textit{SOPInstanceUID}. 
After this, continuous and categorical DICOM tags were separated, and categorical variables were further examined. In particular, some of the tags contained natural language, which fell out of the scope of DICOM tag processing. The eliminated tags include the aforementioned \textit{ProtocolName}, \textit{StudyDescription} and \textit{RequestedProcedureDescription}, accompanied by \textit{AdmittingDiagnosesDescription}, \textit{ImageComments}, etc. 
The remaining categorical variables had no more than $\dcmTagCardinalityThreshold$ unique values. 
After this, {\dcmTagNrOfParsedVars} tags remained, of which {\dcmTagNrOfParsedContinuousVars} were continuous and {\dcmTagNrOfParsedCategoricalVars} were categorical variables.

The final problem to solve regarding DICOM tags was \textbf{missing data analysis}. As was mentioned before, BPE can be directly imputed from other tags via regular expressions, but other values' imputation is not as straightforward. Before imputing these values, the DICOM tags with missing data were analysed further. To determine if data were missing-completely-at-random (MCAR) or missing-at-random (MAR)~\cite{bhaskaran2014differenceMCAR, Emmanuel2021missing_data_ml},  univariate statistical tests were performed as described by Enders~\cite{enders2022applied-missing-data}. Statistical tests differed based on whether the observed variable was discrete or continuous and, in the latter case, if it was normally distributed. If a continuous variable was normally distributed (Shapiro-Wilk, $p\geq0.05$), then an Independent t-test was performed, while a Mann-Whitney U was applied otherwise. In the case of categorical data, a $\chi^{2}$ (chi-square) test was used. A variable would not be considered MCAR if its missingness influenced the distribution of at least one other variable, i.e., there was a statistical difference in the distribution where said variable was missing versus where it was not.
Although the used approach has its drawbacks~\cite{enders2022applied-missing-data}, it can bring attention to dependencies between variables. Furthermore, consulting with the DICOM standard~\cite{dicom_standard, mildenberger2002introduction} strengthens the assumption that data is not MCAR.
The missing values were imputed using MissForest~\cite{Emmanuel2021missing_data_ml, stekhoven2012missforest}, which had been previously shown to work well with MAR data \cite{Tang2017RF_Missingdata}. MSE was used as the criterion for continuous and Gini impurity for categorical variables. After imputation, the categorical variables were one-hot encoded, and continuous variables were scaled to fit the range $[0.00, 1.00]$. We performed a small-scale experiment to observe the influence of MissForest imputation versus simple imputation with the mean value \cite{napravnik2022autoencoders} and observed better scores related to anatomical region grouping.

\section{Image export from DICOM}~\label{AP:DICOM EXPORT}
The material presented here contains the details that complement the text in section~\ref{sec:imgs}.
In order to export images from DICOM files, several parameters are required to be present among the DICOM tags. The initial parameters/tags that require verification are the image (\textit{PixelData} DICOM tag) and modality (\textit{Modality} DICOM tag). Namely, some modalities demand scaling of the raw image pixels' values to values that are meaningful to the targeted application (emphasizing the region of interest for the radiologist). Scaling of each raw image values $x_{Ir}$ is performed using the expression $x_I'= R_s \cdot x_{Ir} + R_i$, where $x_{Ir}$ and $x_I'$ are raw pixel values (the input image) and the rescaled pixel values, respectively; $R_s$ is the rescale slope, and $R_i$ is the rescale intercept. \textit{RescaleSlope} and \textit{RescaleIntercept} are DICOM tags located in the DICOM metadata~\cite{dicom_standard}. 

If DICOM does not contain values for $R_s$ and $R_i$, the default values $R_s = 1$ and $R_i = 0$ are used. In situations where the transformation was not linear, we did not use the look-up table (LUT) for transformation~\cite{thompson:2002:LUT}. We would drop the DICOM in these cases because implementing LUT is case-dependent. The number of cases with LUT in our dataset was negligible, and we could afford to drop the data.

Next, pixel values must be mapped to 8-bit resolution, i.e., interval $[0,255]$. The selection of which values will correspond to $0$ and which to $255$ relies on the specific modality and the organ/tissue being examined. To correctly map pixel values to the 8-bit range, we have followed established practices in radiology which can be formulated as follows \cite{Hrzic:2022:export}:

\begin{equation}
x_I = 
\begin{cases}
  0, & \text{if } x_I' \le W_l \\
  255, & \text{if } x_I' \ge W_u\\
  \frac{1}{W_w} (x_I' - W_c + \frac{1}{2}W_w) \cdot 255 , & \text{otherwise}\\
\end{cases}
\label{eq:transform}
\end{equation}
where $W_l$ and $W_u$ are lower and upper window boundaries, respectively; $\hat{x}_I'$ is the image received after applying rescale slope and intercept; and $x_I$ is the exported 8-bit image. $W_l$ and $W_u$ are calculated as:

\begin{equation}
W_l = W_c - \frac{1}{2}W_w, \hspace{1cm}
W_u = W_c + \frac{1}{2}W_w.
\end{equation}

\noindent
The parameters \textit{WindowCenter} ($W_c$) and \textit{WindowWidth} ($W_w$) are read out from the DICOM metadata. If multiple values were supplied for \textit{WindowCenter} and \textit{WindowWidth}, as is sometimes the case in modalities such as XA, only the first valid value was used. The windowing parameters were considered valid if the resulting image was not entirely monochrome. Also, since some imaging techniques like MR can have multiple slices, we decided only to use the first slice, which contained meaningful data. The meaningful data was selected on two introduced policies.

 The \textit{value policy} was introduced to filter out images containing little-to-no information. \textit{Value policy} calculates the ratio $r_V$ between the number of different pixel values in the image and the total possible number of different pixel values in an image. The calculated ratio $r_V$ was required to be higher than $t_V = 0.1$. Threshold $t_V$ was experimentally determined. 

On the other hand, the \textit{shape policy} demanded that the ratio $r_S$ between exported image width and height was higher than threshold $t_S = 0.1$. Threshold $t_S$ was experimentally defined. By applying this \textit{shape policy}, all erroneous shapes (such as vectors) were filtered out.

To summarise, the raw pixel data is first transferred to the desired pixels' intensity range with the \textit{RescaleSlope} and \textit{RescaleIntercept} parameters. Then, the intensity range of the pixels needs to be converted to the 8-bit range, which is accomplished by utilizing the \textit{WindowCenter} and \textit{WindowWidth} parameters. Finally, it is necessary to check if the content is informative for certain modalities. This is achieved by applying \textit{value} and \textit{shape} policies.

\end{document}